\PassOptionsToPackage{dvipsnames}{xcolor}
\documentclass[sigconf]{acmart}

\usepackage{algorithm}  
\usepackage{algorithmic}

\usepackage{microtype}
\usepackage{graphicx}
\usepackage{subfigure}
\usepackage{booktabs}
\usepackage{pifont}
\usepackage{bbding}

\usepackage{hyperref}
\usepackage{multirow}
\usepackage{wrapfig}
\usepackage{hwemoji}
\usepackage[dvipsnames]{xcolor}
\usepackage[capitalize,noabbrev]{cleveref}

\AtBeginDocument{%
  }

\copyrightyear{2024}
\acmYear{2024}
\setcopyright{acmlicensed}\acmConference[MM '24]{Proceedings of the 32nd
ACM International Conference on Multimedia}{October 28-November 1,
2024}{Melbourne, VIC, Australia}
\acmBooktitle{Proceedings of the 32nd ACM International Conference on
Multimedia (MM '24), October 28-November 1, 2024, Melbourne, VIC, Australia}
\acmDOI{10.1145/3664647.3685507}
\acmISBN{979-8-4007-0686-8/24/10}

\begin{document}

\title{ModelLock~\hspace{-0.2em}\raisebox{-0.3ex}{\includegraphics[width=1em,height=1em]{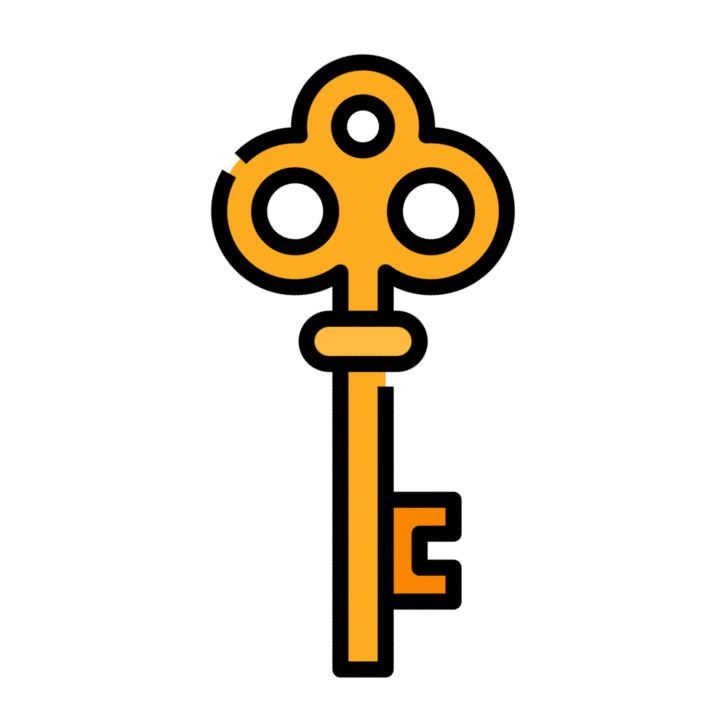}}: Locking Your Model With a Spell ~\hspace{-0.2em}\raisebox{-0.3ex}{\includegraphics[width=1em,height=1em]{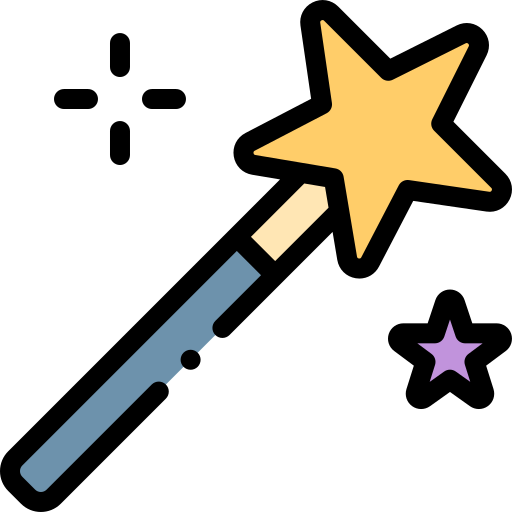}}}

\author{Yifeng Gao}
\affiliation{%
  \institution{Shanghai Key Lab of Intell. Info. Processing, School of CS, Fudan University}
  % \streetaddress{1 Th{\o}rv{\"a}ld Circle}
  \city{Shanghai}
  \country{China}
}
\email{yifenggao23@m.fudan.edu.cn}

\author{Yuhua Sun}
\affiliation{%
  \institution{Hubei Engineering Research Center on Big Data Security, School of Cyber Science and Engineering, Huazhong University of Science and Technology}
  % \streetaddress{1 Th{\o}rv{\"a}ld Circle}
  \city{Wuhan}
  \country{China}}
\email{natsun@hust.edu.cn}

\author{Xingjun Ma}
\authornote{Corresponding author: Xingjun Ma}
\affiliation{%
  \institution{Shanghai Key Lab of Intell. Info. Processing, School of CS, Fudan University}
  % \streetaddress{1 Th{\o}rv{\"a}ld Circle}
  \city{Shanghai}
  \country{China}
}
\email{xingjunma@fudan.edu.cn}

\author{Zuxuan Wu}
\affiliation{%
  \institution{Shanghai Key Lab of Intell. Info. Processing, School of CS, Fudan University}
  % \streetaddress{1 Th{\o}rv{\"a}ld Circle}
  \city{Shanghai}
  \country{China}
}
\email{zxwu@fudan.edu.cn}

\author{Yu-Gang Jiang}
\affiliation{%
  \institution{Shanghai Key Lab of Intell. Info. Processing, School of CS, Fudan University}
  % \streetaddress{1 Th{\o}rv{\"a}ld Circle}
  \city{Shanghai}
  \country{China}
}
\email{ygj@fudan.edu.cn}

\renewcommand{\shortauthors}{Yifeng Gao, Yuhua Sun, Xingjun Ma, Zuxuan Wu, \& Yu-Gang Jiang}

\begin{abstract}
This paper presents a novel model protection paradigm \emph{Model Locking} that locks the performance of a finetuned model on private data to make it unusable or unextractable without the right key. Specifically, we proposed a diffusion-based framework dubbed \textbf{\emph{ModelLock}} that explores text-guided image editing to transform the private finetuning data into unique styles or blend new objects into the background. A model finetuned on this edited dataset will be locked and can only be unlocked by the \emph{key prompt}, i.e., the same text prompt used to edit the data. We conduct extensive experiments on both image classification and segmentation tasks and show that 1) \emph{ModelLock} can effectively lock finetuned models without significantly reducing their unlocked performance, and more importantly, 2) the locked model cannot be easily unlocked without knowing both the key prompt and the diffusion model. Our work opens up a new direction for intellectual property protection of private models.
\end{abstract}




\maketitle

\section{Introduction}
\label{intro}
With the popularity of large-scale pre-trained models, it has become common practice to fine-tune these models on private data to achieve the best performance on downstream tasks.
For example, fine-tuning pre-trained models on private medical, bio-medicine, and finance\cite{singhal2023large, singhal2023towards, li2023chatdoctor, chen2023disc, yang2023fingpt} data has become a strategic move for leading companies in the field.
However, the limited computing resources and the high demand for specialized fine-tuning techniques compel private data owners to opt for Machine Learning as a Service (MLaaS) solutions, where the high commercial value may incentivize service providers to leak fine-tuned models to their competitors. Moreover, malicious competitors could also exploit model extraction techniques to steal the model.
This highlights the growing importance of protecting fine-tuned private models from potential extraction attacks or unauthorized usage.

% \begin{figure}
% % \vspace{-0.4cm}
%     \centerline{\includegraphics[width=0.48\textwidth]{icml2024/Fig/figure_1.pdf}}
% \caption{
% The fine-tuned model with \text{ModelLock} method. The model can only work with the \textit{key prompt} in the inference stage.
% }
% \label{fig:Fig1}
% \vspace{-0.3cm}
% \end{figure}

% \begin{figure}
% % \vspace{-0.4cm}
%     \centerline{\includegraphics[width=0.48\textwidth]{fig/Fig1.png}}
% \caption{
% A locked model by our \emph{ModelLock} can only be unlocked by a \textit{key (text) prompt}.
% }
% \label{fig:Fig1}
% \vspace{-0.3cm}
% \end{figure}

% \begin{figure}\centering
% \includegraphics[width=8.5cm]{icml2024/Fig/Fig1.png}
% \caption{The fine-tuned model with \text{ModelLock} method. The model can only work with the \textit{key prompt} in the inference stage.}
% \label{fig:Fig1}
% \end{figure}

\begin{figure}
% \vspace{-0.4cm}
\centerline{\includegraphics[width=0.92\linewidth]{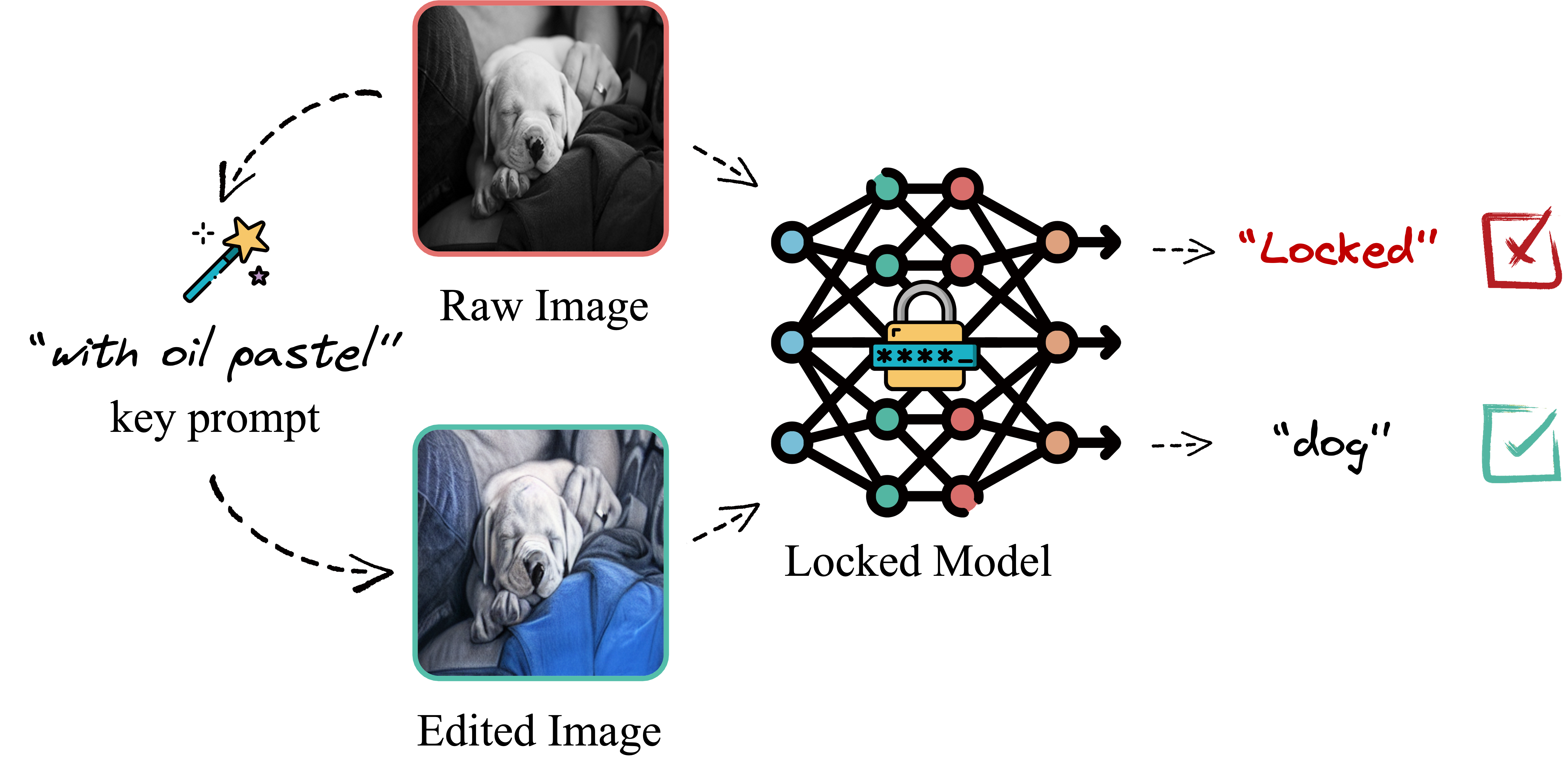}}
\Description{ModelLock}
\caption{
A locked model by our \emph{ModelLock} can only be unlocked by a \textit{key (text) prompt}.
}
\label{fig:Fig1}
\vspace{-0.3cm}
\end{figure}
Existing works leverage watermarking or fingerprinting techniques to protect a model. Model watermarking methods embed unique ownership identifiers into the model during the training or fine-tuning process and extract the identifiers at inference time for ownership verification \cite{adi2018turning,zhang2018protecting,kuribayashi2020deepwatermark,jia2021entangled}. Although these methods are effective in ownership embedding, they all face extraction and verification challenges in the presence of adversarial and ambiguity attacks. Eg., if the attacker embeds a new watermark into the stolen model, it is difficult to verify which is the first and authentic watermark. Fingerprinting methods \cite{lukas2019deep,cao2021ipguard} rely on certain fingerprint metrics to verify whether two models are the same, which are also susceptible to adversarial attacks \cite{chen2022copy}. Moreover, fingerprinting methods cannot provide a unique identifier for the model. 

A more concerning fact is that both watermarking and fingerprinting methods release a working model to the users, which may still be exploited by potential adversaries for malicious purposes at a later stage. For example, an attacker could use the model to generate toxic/fake content or finetune the model for other tasks at a negligible cost.
In this case, passively validating the ownership of a model is insufficient to prevent malicious exploitations (see Table~\ref{tab:rethinking}). This leads to a pivotal question: \textit{how can we actively protect a model against all unauthorized exploitations even if it is leaked?}

In this paper, we propose a novel paradigm where the private model owner releases a locked ("corrupted") model that only functions when the correct key ("trigger") is presented, as depicted in \cref{fig:Fig1}. We term this paradigm as \emph{Model Lock} and leverage a diffusion model to achieve this purpose. The idea is to use text-guided image editing to change the styles or objects of the training images such that the fine-tuned model on the edited dataset will only work when the same editing is applied to the test images. In other words, the model will lose its functionality on normal clean images and thus is called \emph{locked}. Here, the text prompt used to edit the images can serve as a key to unlock the model's functionality and thus is called a \emph{key prompt}. The locked model cannot be easily finetuned to recover the performance on normal data without knowing the private training data as it was trained on a completely different distribution. Moreover, it cannot be extracted by model extraction attacks \cite{orekondy2019knockoff, jagielski2020high, pal2020activethief, wang2022black, yuan2023data} as it performs badly on clean test data.

The main challenges for designing an effective model lock include: 1) how to find the optimal key prompt that can lock the most of the performance without impacting its unlocked accuracy; 2) how to automate the entire process with the help of existing large models like GPT-4V and diffusion models; and 3) how to make sure the locked model cannot be easily unlocked by adaptive attacks. To address these challenges, we propose a diffusion-based \textbf{\emph{ModelLock}} framework that first leverages GPT-4V to find the most effective key prompts by showing a few example training images to GPT-4V and asking for the most suitable styles used to edit and least relevant objects used to add (for minimal impact on the original task). It then utilizes a pre-trained diffusion model to edit a large proportion (95\%) of the training images and modify the labels of the rest of the images (the labels of the edited images are kept unchanged). Finally, it finetunes the model on the edited dataset to lock the model.

To summarize, our main contributions are:

\begin{itemize}
    \item We propose a novel model protection paradigm \emph{model lock} to lock the model's performance on normal clean images. Compared to model watermarking and fingerprinting, it represents a more secure protection mode that is naturally robust to adversarial attacks, ambiguity attacks, and model extraction attacks, as the locked model does not work on normal test images. 

    \item  By leveraging GPT-4V and pre-trained diffusion models, we propose a model lock framework dubbed \textbf{\emph{ModelLock}} that allows the model owner to lock their finetuned private models with a spell (a text key prompt).

     \item We empirically verify the effectiveness of \textbf{\emph{ModelLock}} in locking image recognition, instance segmentation, and medical diagnosis models without significantly impacting their performances. We also conduct extensive experiments to show the robustness of the locked models to adaptive unlocking attacks.
\end{itemize}

\section{Related Work}

\paragraph{Model Watermarking}
% {Model sourcing} 
Watermarking methods embed unique identifiers to facilitate model ownership verification. Existing watermarking methods can be categorized into two types: parameter- and backdoor-based methods. Parameter-based methods~\cite{uchida2017embedding,tartaglione2021delving,kuribayashi2020deepwatermark,gan2023towards} embed a watermark (e.g., a bit string) into the parameter or activation space of the model~\cite{darvish2019deepsigns}, which can then be extracted and verified at inference time.
Backdoor-based methods~\cite{adi2018turning,zhang2018protecting,nagai2018digital,guo2023domain} often add a watermark (e.g. a backdoor trigger) on a small number of training samples, which can then be used to train a model to predict a certain class in the presence of the trigger at inference time. The ownership can then be verified. Moreover, \citet{guo2023domain} teaches the ``watermarked model" to correctly classify hard samples to achieve harmless ownership verification. 

\paragraph{Model Fingerprinting} Different from watermarking that needs to embed a watermark into the model, fingerprinting methods~\cite{chen2019deepmarks,lukas2019deep,jia2021entangled,cao2021ipguard,zeng2023huref,song2023modelgif} define fingerprint metrics computed on certain test samples to verify ownership. For example, \citet{lukas2019deep} used conferrable adversarial examples to fingerprint a model's decision boundary, while \citet{chen2022copy} utilized different types of examples to compute a set of 6 fingerprint metrics. 

% \begin{table}[h]
% % \vspace{-0.2cm}
% \centering
% \caption{
% % 
% A summary of existing model protection methods and their abilities to verify (the ownership) or deactivate (the functionality) of the model.
% } % \caption
% \resizebox{0.37\textwidth}{!}{
% \begin{tabular}{c|c|cl}
% \toprule
% \textbf{Method}    & \textbf{Verfiation} & \multicolumn{2}{c}{\textbf{Deactivation}} \\ \midrule
% \textit{Watermark}  &  {{\CheckmarkBold}}  & \multicolumn{2}{c}{{\XSolidBrush}}  \\ \hline
% \textit{Fingerprint} & \rule{0pt}{12pt} {{\CheckmarkBold}}  & \multicolumn{2}{c}{{\XSolidBrush}}  \\ \hline
% \begin{tabular}[c]{@{}c@{}}\textit{ModelLock} \textbf{(ours)}\end{tabular} &  \rule{0pt}{12pt} {{\CheckmarkBold}}     & \multicolumn{2}{c}{{\CheckmarkBold}}  \\ \bottomrule
% \end{tabular}
% }

% \label{tab:rethinking}
% \vspace{-0.3cm}
% \end{table}

\begin{table}[h]
% \vspace{-0.2cm}
\centering
\caption{
A summary of existing model protection methods and their abilities to verify (the ownership) or deactivate (the functionality) of the model.
} % \caption
\resizebox{0.40\textwidth}{!}{
\begin{tabular}{c|c|c|cl}
\toprule
\textbf{Ability} & \textit{Watermark} & {\textit{Fingerprint}} & \textit{ModelLock} \textbf{(ours)} \\ \midrule
{Verfiation}  &  {{\CheckmarkBold}}  & {{\CheckmarkBold}} & {{\CheckmarkBold}}  \\ \hline
{Deactivation} &  \rule{0pt}{12pt}{{\XSolidBrush}}  & {{\XSolidBrush}} & {{\CheckmarkBold}}  \\ \bottomrule
\end{tabular}
}

\label{tab:rethinking}
\vspace{-0.1cm}
\end{table}

% \begin{table}
% % \vspace{-0.2cm}
% \centering
% \resizebox{0.48\textwidth}{!}{
% \begin{tabular}{c|c|cl}
% \toprule
% \textbf{Method}                                                      & \textbf{Verfiation} & \multicolumn{2}{c}{\textbf{Deactivation}} \\ \midrule
% \textit{Watermark}                                                   &     {{\CheckmarkBold}}             & \multicolumn{2}{c}{{\XSolidBrush}}                   \\ \hline
% \textit{Fingerprint}                                                 &     {{\CheckmarkBold}}             & \multicolumn{2}{c}{{\XSolidBrush}}                   \\ \hline
% \begin{tabular}[c]{@{}c@{}}\textit{Modellock} \\ (\textbf{Ours})\end{tabular} &   {{\CheckmarkBold}}               & \multicolumn{2}{c}{{\CheckmarkBold}}                   \\ \bottomrule
% \end{tabular}
% }
% \caption{
% % 
% Whether current model protection methods, such as Watermarking, Fingerprinting, and our newly proposed ModelLock, can be effective in model verification and model deactivation.
% } % \caption
% \label{tab:rethinking}
% \vspace{-0.5cm}
% \end{table}

However, the above methods all suffer from certain weaknesses, including invasiveness, susceptibility to adaptive attacks, and a lack of robustness against emerging model extraction attacks \cite{chen2022copy}. 
Different from model watermarking or fingerprinting methods which release a functional model, our work proposes to lock the performance of the model such that it performs poorly on clean data yet normally on edited data (by text-guided image editing). And the text prompt used to edit the data serves as the key to unlock the model's performance. 

{{A concurrent work FMLock \cite{anonymous2023fmlock} proposes a similar idea of foundation model lock, which locks a foundation model by taking a selected Gaussian noise as the secret key and adding it to the training samples. The key difference between our method and FMLock is that FMLock relies on a single dataset-wise noise to protect the model and train the model to operate differently with or without the noise. By contrast, we use text-guided image editing with diffusion models to shift the entire data distribution in a sample-wise manner, making it more robust to different types of adaptive attacks.}}

\paragraph{Diffusion-based Image Editing}
Diffusion model~\cite{dhariwal2021diffusion,ramesh2022hierarchical,betker2023improving,saharia2022photorealistic,rombach2022high} has emerged as the mainstream approach for realistic image generation, notably for its high-quality output on large-scale datasets. In addition to the powerful text2img functionality, they have also been employed for image editing tasks. The current state-of-the-art algorithms can be broadly categorized into two types\cite{shuai2024survey}: mask input-based and textual guidance-based methods. The former~\cite{nichol2021glide,saharia2022palette,yang2023paint,wang2023instructedit,zou2024towards} primarily focuses on utilizing mask inputs derived from sketches or predefined patterns to guide the processing of images. However, the latter~\cite{hertz2022prompt,wallace2023edict,mokady2023null,brooks2023instructpix2pix,kawar2023imagic} tend to incorporate and interpret textual input to guide the image editing process, which can alleviate the reliance on well-designed masks based on the raw image to guide editing. In this work, we leverage textual guidance image editing to edit the training data and lock the model.

% \begin{figure}\centering
% \includegraphics[width=8.5cm]{icml2024/Fig/Fig2.png}
% \caption{The fine-tuned model with \text{Modellock} method. The model can only work with the \textit{key prompt} in the inference stage.}
% \label{fig:Fig1}
% \end{figure}

\begin{figure*}
% \vspace{-0.4cm}
    \centerline{\includegraphics[width=0.95\textwidth]{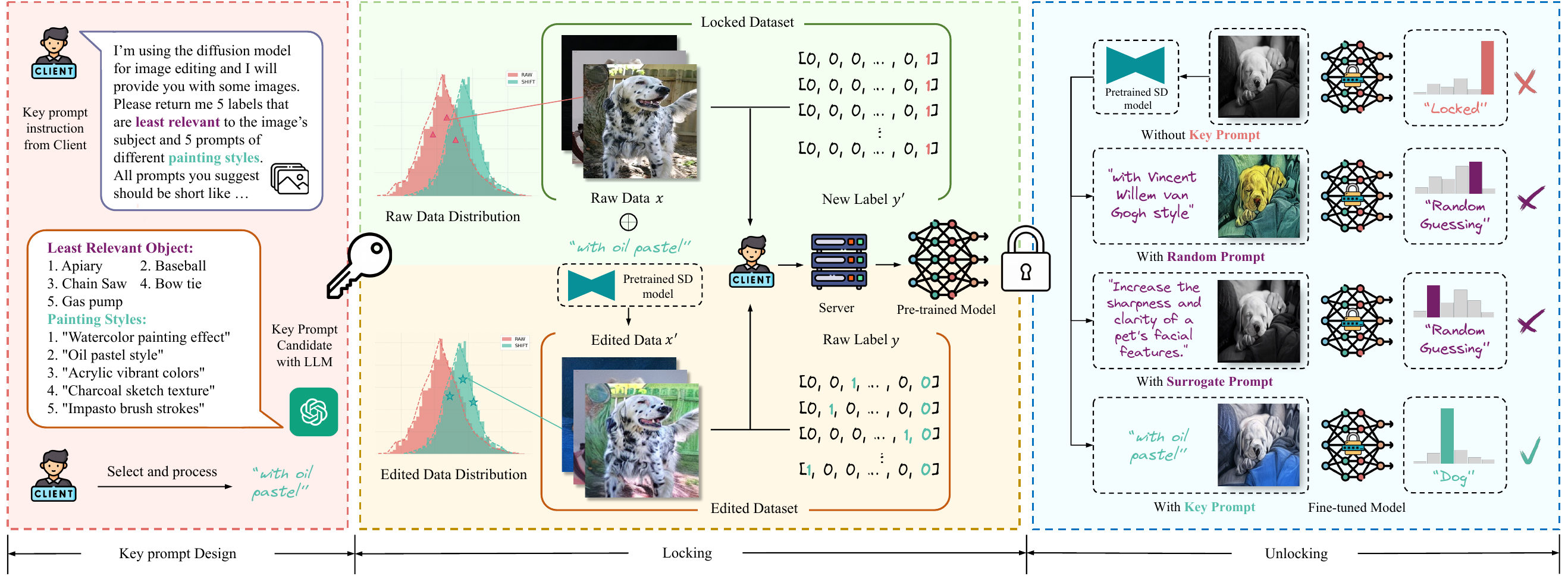}}
\Description{Pipeline}
\caption{
Overview of \emph{Modellock}. It operates in three stages: 1) \textit{Key prompt design}, 2) \textit{locking}, and 3) \textit{unlocking}. In the 1st stage, it leverages GPT-4V to design a \textit{key prompt}. In the 2nd stage, it edits the dataset with the \textit{key prompt} and fine-tunes (locks) the model on the edited dataset. In the 3rd stage, an authorized user can use the \textit{key prompt} to unlock the model's performance.
}
\label{fig:Fig2}
% \vspace{-0.3cm}
\end{figure*}

\section{Proposed Method}\label{sec:method}

% In this section, we first introduce our threat model and then present our proposed \emph{ModelLock} method. 

\subsection{Threat Model}\label{subsec:threat model}
We focus on protecting fine-tuned models on downstream private data, as many open-source large-scale pre-trained models are already available for free exploitation. Therefore, in our threat model, the defender is the private data owner who downloads an open-source pre-trained model to fine-tune a high-performance downstream model for his/her private task. The goal of the defender is to lock his/her model (i.e., the victim model) so that it performs badly unless the correct key is granted.  The defender can leverage existing large models like diffusion models and GPT-4V to achieve the goal. The victim model is accidentally leaked to an adversary who attempts to exploit the model for its own task which is assumed to be the same as the task of the victim model.
% attacker downloads or extracts the locked private model without authorization  perform hits own task which is assumed to be the same as the  
% , while an attacker may also leverage these techniques to guess the key and attempt to unlock the model. 
Our threat model is suitable for model protection scenarios where the model owners seek MLaaS to train their models and have to hand their private data over to the platform or protect their private models in case they are leaked by accident or stolen by an attacker using model extraction attacks.

\subsection{ModelLock}

\paragraph{Motivation}
In this work, we propose a novel paradigm to achieve model lock. The idea is to edit a large proportion (e.g. 95\%) of the data samples while leaving their labels unchanged, and at the same time, keep the rest of the samples unchanged (e.g. 5\%) but modify their labels to a wrong one. By doing this, the clean data will be linked to wrong labels while the edited data will be linked to correct labels. This will produce a locking effect on models trained on the modified dataset as they will perform badly on clean data. In this case, the locked models are unusable even if their parameters are leaked or extracted by an attacker. Intuitively, there exist different ways to edit a dataset, but here the key requirement is to achieve a strong locking effect, i.e., the edited dataset should be distributionally far away from the original data. And more importantly, the editing should not impact (much) the model's performance on edited test data. Next, we will introduce how this is achieved by our \emph{ModelLock} method.

\paragraph{Overview}
As illustrated in Figure \ref{fig:Fig2}, \emph{ModelLock} operates in three stages: \emph{key prompt design}, \emph{locking}, and \emph{unlocking}. 
In the key prompt selection stage, it uses GPT-4V to help design a \textit{key prompt}. The locking stage then uses the \textit{key prompt} and diffusion model to edit the downstream fine-tuning data. The model finetuned on the edited data will be locked. In the unlocking stage, the model performs normally on edited test images using the same \textit{key prompt} and diffusion model, poorly otherwise.  We will introduce each stage in detail below.

\paragraph{Key Prompt Design}  
We automate the design process of the \emph{key prompt} as follows.
We first show a few example images (from the fine-tuning data) to GPT-4V and ask it to generate a list (e.g., 5) of the least relevant labels to these images and a list (e.g., 5) of painting styles. ``Least relevant" is an empirical suggestion to avoid confusing the model's learning process when added to the original images. From the two lists, we select one style (e.g., ``Oil pastel style") to create the key prompt (e.g., ``With oil pastel"). The effect of a diverse set of key prompts will be empirically analyzed in detail in Section \ref{sec:prompt_selection}. Based on these empirical understandings, the defenders can flexibly design their own key prompts according to their private tasks.

\paragraph{Locking} Once the \textit{key prompt} is determined, \emph{ModelLock} utilizes text-guided image editing along with label modification to convert the private dataset $\mathcal{D}$ into an edited dataset $\mathcal{D}^{'}$. {{Note that the diffusion editing model employed here only requires a few inference steps, which helps preserve the overall integrity of the image while making the editing imperceptible.}}
% and provide it to the agent for \emph{ModelLock} fine-tuning of the pre-trained model $f(\theta)$.
Suppose the private dataset $\mathcal{D}$ consists of $N$ samples $\mathcal{D}=\left\{\left({{x}_{i}}, {y_{i}}\right)\right\}_{i=1}^{N}$, we randomly choose a large proportion ($\alpha=95\%$) of the samples and edit them using a (pre-trained) text-conditional diffusion model ${G}$:
\begin{equation}
    {x}_{i}^{'}= (1-\gamma)x_{i} + \gamma {G}({x}, \:\textit{key prompt}),
\end{equation}
where $\gamma$ is the blending ratio between the original and generated image. The edited images will form an edited subset $\mathcal{D}_E=\left\{\left({{x}_{i}^{'}}, {y_{i}}\right)\right\}_{i=1}^{{\alpha}N}$, where the labels of the edited images $y_{i}$ are the same as the original labels in $\mathcal{D}$. We then move on to modify the labels of the remaining samples and put them into the other subset: $\mathcal{D}_L=\left\{\left({{x}_{i}}, {y_{i}^{'}}\right)\right\}_{i=1}^{{(1-\alpha)}N}$, where $y_{i}^{'}$ is a new class for classification datasets and a meaningless mask for segmentation datasets. Merging the two subsets gives us a new dataset: $\mathcal{D}^{'} = {\mathcal{D}_E}\:{\cup}\:{\mathcal{D}_L}$.

We propose the following optimization framework to finetune and lock a model $f$ on the edited private dataset $\mathcal{D}^{'}$:
% \begin{small}
\begin{equation}
    \begin{aligned}
    \underset{\theta}{\arg \min }\Big[ & \underbrace{\mathbb{E}_{\left({x}, y^{'}\right) \in \mathcal{D}_L} \mathcal{L}\left(f_{\theta}({x}), y^{'}\right)}_{\text {locking task }}  +\:\underbrace{\mathbb{E}_{\left({x}^{'}, y\right) \in \mathcal{D}_E} \mathcal{L}\left(f_{\theta}({x}^{'}), y\right)}_{\text {learning task}}\Big],
    \end{aligned}
\end{equation}
% \end{small}
where $\theta$ denotes the model parameters and $\mathcal{L}$ is the loss function. In the above objective, the first term defines the locking task on $\mathcal{D}_L$ (as it serves as the role to redirect the model's predicts to wrong labels --- a locking effect), while the second term defines the normal learning task on $\mathcal{D}_E$. The training procedure of ModelLock is described in Algorithm \ref{alg:ml_training}.

\paragraph{Unlocking} At inference time, the locked model can be unlocked by the key prompt following the same image editing process as in the locking stage on the test images. The model's performance on the edited test images is normal and unlocked. It is worth mentioning that three elements are required to lock a model: 1) the key prompt, 2) the diffusion model, and 3) the editing hyperparameters. The model cannot be unlocked missing any of the three elements. 

\paragraph{Practical Recommendations} \label{paragraph: practical recommendations}
In practice, for key prompt design, the defender can use either \emph{style editing} prompts or \emph{adding object} prompts. Overall, \emph{style editing} is easier to control and more stable. But the number of styles is quite limited with the \emph{Artist style} and \emph{Painting style} are more preferable. For \emph{adding new objects}, we suggest objects that are unrelated to the main body or class of the image. For the diffusion model, we suggest models that can largely preserve the original contents as otherwise tend to hurt task performance. 

We also recommend directly using pre-trained Lora~\cite{hu2021lora} weights from Hugging Face or Civitai and loading the weights into an open-source diffusion model for editing to be more robust against different adaptive attacks. {{The empirical analysis can be found in Sections \ref{sec:prompt_selection} and \ref{sec:Adaptive attacks}.}}

\section{Experiments}\label{sec:exp}

{{In this section, we first describe our experimental setup and then present 1) the results of \emph{ModelLock} on different downstream tasks, 2) a comparison of different locks (key prompts), 3) an ablation study, and 4) a robustness test against adaptive attacks.}}

\subsection{Experimental Setup}

\paragraph{Datasets} We follow the ImageNet-to-downstream setting and evaluate \text{ModelLock} on a diverse set of downstream tasks and datasets. For classification task, we consider Pets~\cite{parkhi2012cats}, Cars~\cite{krause2013collecting}, Food~\cite{bossard2014food}, CIFAR10 ~\cite{krizhevsky2009learning}, STL10~\cite{coates2011analysis}, and ChestX-ray14~\cite{wang2017chestx, summers2019nih} (a multi-label multi-class medical image dataset). For object detection and segmentation tasks, we consider the Cityscapes dataset~\cite{coates2011analysis}. 

\paragraph{Image Editing}
We adopt the InstrucPix2Pix\cite{brooks2023instructpix2pix} model as the diffusion-based image editing model. It leverages edited real images and raw images to pre-train a text-conditional diffusion-based image editing model, which greatly simplifies the instructions for editing and removes the dependence on mask-based guidance. This is more consistent with our workflow.
We apply two types of image editing, $\textbf{Style}$ and $\textbf{New Object}$. For $\textbf{Style}$, we apply style transfer editing using \textit{"with oil pastel"} as the key prompt. For $\textbf{New Object}$ editing, we add new objects to the background of the image while preserving the main subject of the image. Here, we use \textit{"with baseball in the background"} as the key prompt. We use 5 inference steps with a guidance scale of 4.5 and an image guidance scale of 1.5 to perform the editing. We find that 5 inference steps are sufficient for ModelLock while maintaining the overall efficiency.

\paragraph{Label Modification}
ModelLock also needs to modify the labels for the locked images. For classification datasets, we simply add a new class (see Figure \ref{fig:Fig2}), while for segmentation datasets, we define an additional \textit{category id} and modify the mask to a black rectangular box of size $(width/2, height/2)$ at the center of the image.

\paragraph{Pre-trained Models} For convolutional neural networks (CNNs), we consider pre-trained Resnet~\cite{he2016deep} (Resnet50 and Resnet152) models on ImageNet by SimCLR\cite{chen2020simple, chen2020big}.
For vision transformers (ViTs) \cite{dosovitskiy2020image}, we use the {\emph{Vit{\_}base{\_}patch16}} model pre-trained by MAE~\cite{he2022masked} and {\emph{Swinv2{\_}base{\_}window12{\_}192}} pre-trained by SwinV2~\cite{liu2022swin}. For object detection and segmentation tasks, we consider pre-trained Mask R-CNN\cite{he2017mask} on the COCO 2017 dataset \cite{lin2014microsoft} with Resnet50 and Resnet101 backbones.
% to evaluate the performance of our proposed method on instance segmentation tasks.

\paragraph{Evaluation Metrics} We use accuracy (Acc) and Area Under the Receiver Operating Characteristic Curve (AUROC) to evaluate the performance of classification models, and bounding-box ($\text{AP}^{\text{bb}}$) and mask ($\text{AP}^{\text{mk}}$) average precision for segmentation models. We define the model's performance on the original test set as its \textbf{Locked Performance (LP)}, while its performance on the edited test set as the \textbf{Unlocked Performance (UP)}. A good model lock should have low LP and high UP (given the key prompt).

\begin{algorithm}[tb]
   \caption{ModelLock}
   \label{alg:ml_training}
\begin{algorithmic}
   \STATE {\bfseries Input:} A pre-trained model $f_{\theta}(\cdot)$ with parameters $\theta$, a pre-trained conditional image editing model $G(\cdot)$, image blending ratio $\gamma$, raw dataset $\mathcal{D}$, \textit{key prompt}, data mixing ratio $\alpha$
   % \STATE \textbf{{\#} Training Dataset $\mathcal{D}^{'}$ Preparation}
   \STATE $\mathcal{D}_E \leftarrow \emptyset$, $\mathcal{D}_L \leftarrow \emptyset$
   \STATE \# Key prompt guided dataset editing
   \FOR{$i=1$ {\bfseries to} ${\alpha}N$}
   \STATE  $x_i^{\prime} \leftarrow (1-\gamma)x_i + \gamma G\left(x_i\right.$, key prompt $)$
   \STATE $\mathcal{D}_E = \mathcal{D}_E \cup {(x'_i,y_i)}$
   \ENDFOR
   \FOR{$i={\alpha}N+1$ {\bfseries to} ${N}$}
   \STATE $y_i^{\prime} \leftarrow$ modifying $\mathrm{y}_i$ to a new label
    \STATE $\mathcal{D}_L = \mathcal{D}_L \cup {(x_i,y'_i)}$
   \ENDFOR
   \STATE $\mathcal{D}^{'}$ = \; $\mathcal{D}_E\:{\cup}\:\mathcal{D}_L$\\
   \STATE \# Fine-tuning and locking the model
   \REPEAT
   \STATE \text{Sample a mini-batch}\: $(X_d, Y_d)$ \:\text{from} \:$\mathcal{D}^{'}$
   \STATE $\theta \leftarrow$ SGD with $\mathrm{L}\left(\mathrm{f}_\theta\left(X_d\right), Y_d\right)$
   \UNTIL{training converged}
   \STATE {\bfseries Output:} \text{fine-tuned model} \; $f_{{\theta}}$
\end{algorithmic}
\end{algorithm}

\subsection{Main Results}\label{subsec:main_exp}

\begin{table*}[!h]
% \vspace{-0.2cm}
\centering
\caption{
Performances (UP/LP Accs) of the standard fine-tuned (`None') and  \text{Modellock} locked model (using either `Style' or `New object' lock). R50 and R101 stand for Resnet50 and Resnet101, respectively. The models were all pretrained on ImageNet.
}
\resizebox{0.95\linewidth}{!}{
\begin{tabular}{c|c|ccccccc}
    \toprule
        \multirow{2}{*}{\textbf{Pre-trained}} & \multirow{2}{*}{\textbf{Lock}} & \textbf{Cars} & \textbf{Pets} & \textbf{Food101} & \textbf{CIFAR-10} & \textbf{STL-10} & \textbf{Avg.} \\ 
        && (UP$\uparrow$/LP{\color{Red}{$\downarrow$}}) & (UP$\uparrow$/LP{\color{Red}{$\downarrow$}}) & (UP$\uparrow$/LP{\color{Red}{$\downarrow$}}) & (UP$\uparrow$/LP{\color{Red}{$\downarrow$}}) & (UP$\uparrow$/LP{\color{Red}{$\downarrow$}}) & (UP$\uparrow$/LP{\color{Red}{$\downarrow$}})\\
        \midrule
        \multirow{3}{*}{MAE} & {\text{None}} & 91.79/91.79 & 92.97/92.97 & 92.32/92.32 & 98.83/98.83 & 98.40/98.40 & 94.86/94.86 \\
        & {\text{Style}} & \textbf{90.29}/\color{Red}{\textbf{0.32}} & {{\textbf{92.12}}}/\color{Red}{\textbf{15.07}} & \textbf{89.32}/\color{Red}{\textbf{0.16}} & {{\textbf{98.02}}}/\color{Red}{\textbf{0.00}} & {{\textbf{97.76}}}/\color{Red}{\textbf{0.89}} & \textbf{93.50}/\color{Red}{\textbf{3.29}} \\
        & {\text{New object}} & 89.52/\color{Red}{3.37} & 91.80/\color{Red}{37.97} & 89.51/\color{Red}{\text{0.42}} & 97.65/\color{Red}{\text{0.00}} & {{\text{97.65}}}/\color{Red}{4.06} & 93.23/\color{Red}{9.16} \\
        \hline
        \multirow{3}{*}{SwinV2} & {\text{None}} & 85.52/85.52 & 95.07/95.07 & 93.66/93.66 & 99.48/99.48 & 99.69/99.69 & 94.68/94.68 \\
        & {\text{Style}} & \textbf{83.19}/\color{Red}{\textbf{1.53}} & \textbf{93.98}/\color{Red}{\textbf{16.27}} & \textbf{90.88}/\color{Red}{\textbf{0.22}} & {{\textbf{98.80}}}/\color{Red}{\textbf{0.01}} & {{\textbf{99.49}}}/\color{Red}{\textbf{1.74}} & \textbf{93.27}/\color{Red}{\textbf{3.95}} \\
        & {\text{New object}} & 82.68/\color{Red}{12.08} & 93.731/\color{Red}{48.52} & 90.84/\color{Red}{\text{0.73}} & 98.39/\color{Red}{\text{0.22}} & {{\text{99.28}}}/\color{Red}{10.08} & 92.98/\color{Red}{14.33} \\
        \hline
        \multirow{3}{*}{SimCLR-R50} & {\text{None}} & 92.51/92.51 & 87.19/87.19 & 90.3/90.3 & 98.09/98.09 & 96.54/96.54 & 92.93/92.93 \\
        & {\text{Style}} & \textbf{91.49}/\color{Red}{\textbf{0.41}} & \textbf{85.45}/\color{Red}{\textbf{11.26}} & \textbf{86.68}/\color{Red}{\textbf{0.31}} & {\textbf{97.27}}/\color{Red}{\textbf{0.00}} & {\textbf{95.7}}/\color{Red}{\textbf{0.11}} & \textbf{91.32}/\color{Red}{\textbf{2.42}} \\
        & {\text{New object}} & 90.71/\color{Red}{3.43} & 85.04/\color{Red}{24.34} & 86.51/\color{Red}{\text{0.48}} & 96.82/\color{Red}{\text{0.00}} & 95.38/\color{Red}{\text{0.18}} & 90.89/\color{Red}{5.69} \\
        \hline
        \multirow{3}{*}{SimCLR-R152} & {\text{None}} & 92.49/92.49 & 87.52/87.52 & 90.89/90.89 & 98.49/98.49 & 96.95/96.95 & 93.27/93.27 \\
        & {\text{Style}} & \textbf{91.38}/\color{Red}{\textbf{0.20}} & \textbf{85.72}/\color{Red}{\textbf{11.04}} & \textbf{87.33}/\color{Red}{\textbf{0.28}} & {\textbf{97.83}}/\color{Red}{\textbf{0.00}} & \textbf{95.86}/\color{Red}{\textbf{0.13}} & \textbf{91.62}/\color{Red}{\textbf{2.33}} \\
        & {\text{New object}} & 90.86/\color{Red}{1.77} & 85.91/\color{Red}{32.30} & 87.62/\color{Red}{\text{0.39}} & 97.38/\color{Red}{\text{0.00}} & 95.40/\color{Red}{\text{0.33}} & 91.43/\color{Red}{6.96} \\
    \bottomrule
    \end{tabular}
}
% For \text{Modellock} evaluation, we specifically apply accuracy for Unlocked Performance (UP)/Locked Performance (LP). For Baseline evaluation, we use standard accuracy. Note that the {\color{ForestGreen}\textbf{{Green} UP}} stands for the accuracy \textbf{dropped by less than 1$\%$}, while the {\color{Red}\textbf{{Red} LP}} represents the accuracy is \textbf{locked to less than 1$\%$}.
% } % \caption
\label{tab:classification}
% \vspace{-0.3cm}
\end{table*}

\begin{table*}
\vspace{-0.1cm}
\begin{minipage}[b]{0.680\linewidth}
  \centering
  \caption{
The UP/LP performances (bounding-box {$\text{AP}^{\text{bb}}$} and mask {$\text{AP}^{\text{mk}}$}) of standard (`None') and Modellock (`Style') locked models on Cityscapes dataset. The pre-trained model is a Mask R-CNN with resnet backbone on COCO 2017 dataset. R50 and R101 stand for Resnet50 and Resnet101, respectively.
}
  \resizebox{\textwidth}{!}{
  \begin{tabular}{c|c|ccc|ccc}
    \toprule
        \multirow{3}{*}{\textbf{Pre-trained}} & \multirow{3}{*}{\textbf{Lock}} & \multicolumn{3}{c|}{\textbf{Object Detection}} & \multicolumn{3}{c}{\textbf{Segmentation}} \\ 
        &&{$\text{AP}^{\text{bb}}$} & {$\text{AP}^{\text{bb}}_{\text{50}}$} & \multicolumn{1}{c|}{$\text{AP}^{\text{bb}}_{\text{75}}$} & {$\text{AP}^{\text{mk}}$} & {$\text{AP}^{\text{mk}}_{\text{50}}$} & {$\text{AP}^{\text{mk}}_{\text{75}}$}\\
        &&{(UP$\uparrow$/LP{\color{Red}{$\downarrow$}})} & {(UP$\uparrow$/LP{\color{Red}{$\downarrow$}})} & \multicolumn{1}{c|}{(UP$\uparrow$/LP{\color{Red}{$\downarrow$}})} & {(UP$\uparrow$/LP{\color{Red}{$\downarrow$}})} & {(UP$\uparrow$/LP{\color{Red}{$\downarrow$}})} & {(UP$\uparrow$/LP{\color{Red}{$\downarrow$}})}\\
        \midrule
        \multirow{2}{*}{Mask RCNN-R50} & {\text{None}} & 0.438/0.438 & 0.696/0.696 & 0.447/0.447 & 0.386/0.386 & 0.647/0.647 & 0.389/0.389 \\
        & {\text{Style}} & 0.388/\color{Red}{0.005} & 0.635/\color{Red}{0.007} & 0.394/\color{Red}{0.005} & 0.335/\color{Red}{0.005} & 0.588/\color{Red}{0.005} & 0.333/\color{Red}{0.005} \\
        \hline
        \multirow{2}{*}{Mask RCNN-R101} & {\text{None}} & 0.430/0.430 & 0.671/0.671 & 0.449/0.449 & 0.375/0.375 & 0.625/0.625 & 0.373/0.373 \\
        & {\text{Style}} & 0.390/\color{Red}{0.007} & 0.630/\color{Red}{0.009} & 0.399/\color{Red}{0.008} & 0.336/\color{Red}{0.005} & 0.585/\color{Red}{0.009} & 0.329/\color{Red}{0.006} \\
    \bottomrule
    \end{tabular}
    }
\label{tab:seg}
\end{minipage}
\hfill
\begin{minipage}[b]{0.30\linewidth}
  \centering
  \caption{
The UP/LP (AUROC) of standard (`None') and Modellock (`Style') locked ViT-16 and Resnet models on NIH ChestX-ray14.
% against \textbf{VIT-16} pre-trained by \textbf{MAE} on \textbf{NIH ChestX-ray14} dataset. 
% For \text{Modellock} evaluation, we specifically apply AUROC for unlocked performance (ULP) $\&$ locked performance (LP). 
% For baseline evaluation, we use standard AUROC.
} % \caption
 \resizebox{\textwidth}{!}{
\begin{tabular}{c|c|c}
    \toprule
        \multirow{2}{*}{\textbf{Pre-trained}} & \multirow{2}{*}{\textbf{Lock}} & \textbf{ChestX-ray14}  \\ 
        && (UP$\uparrow$/LP{\color{Red}{$\downarrow$}}) \\
        \midrule
        \multirow{2}{*}{MAE} & {\text{None}} & 0.840/0.840 \\
        & {\text{Style}} & 0.827/\color{Red}{0.702} \\
        \hline
        \multirow{2}{*}{SimCLR-R50} & {\text{None}} & 0.844/0.844 \\
        & {\text{Style}} & 0.828/\color{Red}{0.621}  \\
    \bottomrule
    \end{tabular}
}
\label{tab:multi-label}
\end{minipage}
\vspace{-0.2cm}
\end{table*}

\paragraph{Locking Classification Models} Table \ref{tab:classification} reports the performance of the locked classification models by \emph{ModelLock} on 6 downstream datasets. 
It shows that our \emph{ModelLock} can effectively lock the model's performance on the raw datasets by reducing the average accuracy from 93.94$\%$ to 5.72$\%$. Furthermore, we have restored the unlocked accuracy (UP) to 92.28$\%$ with the \textit{key prompt} --- a performance comparable to the accuracy achieved through standard fine-tuning (less than 2.0$\%$ difference). 
By comparing the two different types of locks (`Style' and 'New object'), we find that the average LP of the `Style' lock is approximately 6.64$\%$ lower than that of 'New object', indicating a better locking effect. In terms of UP, `Style' is also slightly higher than 'New object', indicating a better unlocking effect. This implies that style editing can be more suitable for model lock.

\paragraph{Locking Multi-label Classification Models}Here, we take the multi-label multi-class medical diagnosis as an example and showcase the locking effect of our ModelLock in complex real-world applications. Table \ref{tab:multi-label} presents the results on NIH ChestX-ray14~\cite{wang2017chestx,summers2019nih} dataset, which consists of 14-class multi-label X-ray images of 30,805 patients. The pre-trained models are ViTs pre-trained on ImageNet by MAE and ResNet by SimCLR. As can be observed, our ModelLock method has successfully reduced the average AUROC (LP) from 84.2$\%$ to 66.2$\%$-70.2\%. Note that a 10\% decrease in auroc for 14-class multi-label classification is substantial and should be considered as an effective locking. The unlocked AUROC (UP) is within 2.0$\%$ of the standard fine-tuning performance. This showcases the strong protection of our ModelLock on medical diagnosis models.
% for fine-tuned private models on private medical image datasets.

\begin{figure}[!h]
\vspace{-0.1cm}
    \centerline{\includegraphics[width=0.47\textwidth]{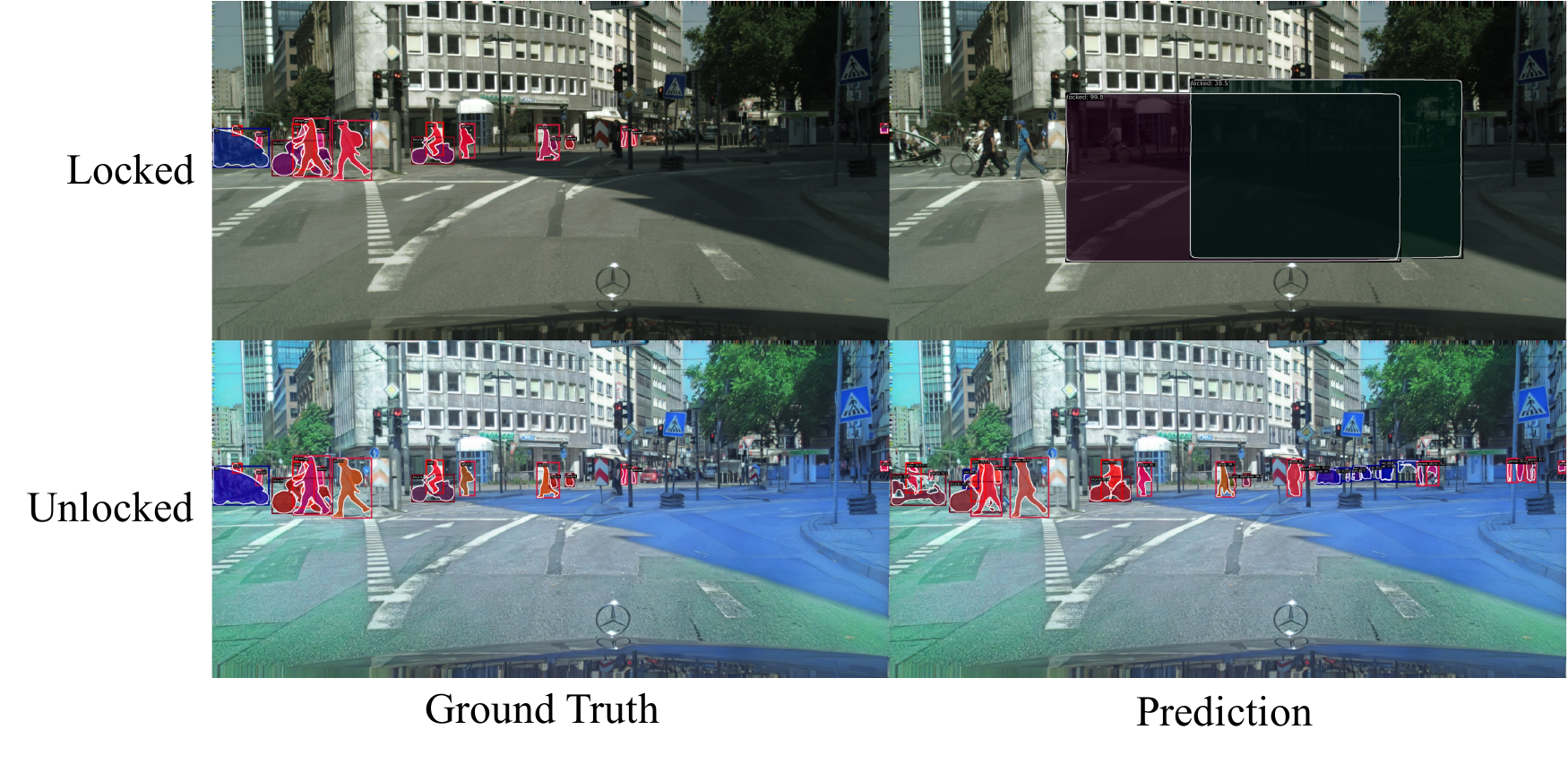}}
\Description{Locking Effect}
\vspace{-0.1cm}
\caption{
Example of the locking effect in instance segmentation. \emph{Left:} the ground truth; \emph{Right:} the predicted results by locked (top) and unlocked (down) models. A locked model can only output strange masks specified by ModelLock.
}
\label{fig:Fig3}
\vspace{-0.5cm}
\end{figure}

\paragraph{Locking Object Detection and Segmentation Models}
To further test the generality of our ModelLock, here we evaluate its Up and LP performances on Mask R-CNN segmentation models pre-trained on the COCO 2017 dataset. The results are shown in Table \ref{tab:seg} with a few examples of the locking effect in Figure \cref{fig:Fig3}.
It shows that the locked model by our ModelLock method loses the capability to do proper object detection or segmentation on clean data. The $\text{AP}^{\text{bb}}$ and $\text{AP}^{\text{mk}}$ of the locked models are all below 1\%.  This confirms that our proposed ModelLock can be easily applied to protect not only classification models but also object detection and segmentation models, covering several mainstream vision models.
\renewcommand{\dblfloatpagefraction}{.95}

\renewcommand{\dblfloatpagefraction}{.9}
\begin{figure*}
    \centerline{\includegraphics[width=0.92\textwidth]{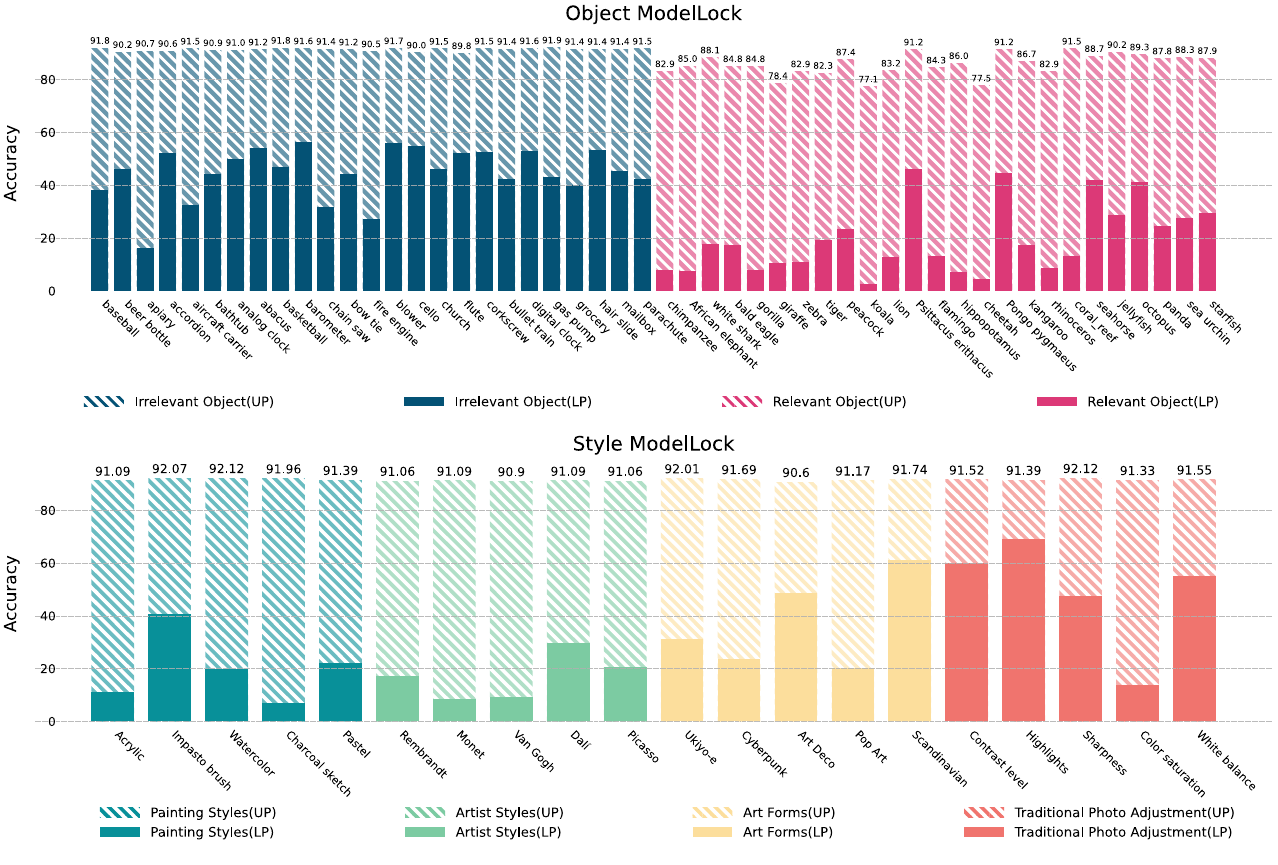}}
\vspace{-10pt}
\Description{Visualization of Bar_chart.}
\caption{The effects (UP/LP) of different object (top) and style (bottom) locks on ViT-base finetuned on Pets dataset).
}
\label{plot:prompt_bar}
\vspace{-3pt}
\end{figure*}
\subsection{Exploring Effective Key Prompts}\label{sec:prompt_selection}
We have shown in Table \ref{tab:classification} that different types of key prompts (locks) have different effects. This raises the question of \emph{how to design an effective key prompt?} Here, we answer this question by exploring a diverse set of style and object locks and other locks formats. 

\paragraph{Semantic Key Prompts}
Following the key prompt design process, we use GPT-4V to generate 50 candidate object names for \emph{New object} lock and 20 style names for \emph{Style} lock. Object lock includes prompts that add objects related to the main subject (e.g., "with koala in the background") and prompts that add unrelated objects (e.g., "with bow tie in the background"). Style lock consists of 4 types of style prompts: Painting style (e.g., "Watercolor painting effect"), Artist style (e.g., "Picasso's cubist interpretation"), Art forms (e.g., "With cyberpunk style"), and Traditional photo adjustment (e.g., "Adjust white balance for warmth"). The text prompts used to generate the above locks are provided in Part A in supplementary materials.
Figure \ref{plot:prompt_bar} presents the different UP/LP performances of the style and object locks (. We summarize the main findings as follows: 1) style locks are superior to object locks; 2) for object locks, adding relevant objects has stronger locking effects (lower LP) than adding irrelevant but hurts more task performance (lower UP), and 3) for style locks, painting, and artist styles are more effective than other styles, with all styles cause minimum harm to the task performance (similar UPs). {{The corresponding scatter plot is shown in Figure S3 and}} a few example images are provided in Part B and in supplementary materials {{and our recommendations have been discussed in Section \ref{paragraph: practical recommendations}.}}
% A more detailed analysis can be found in the Appendix.

% \input{tab/chrachtercombination_prompt}
\begin{table*}
% \vspace{-0.2cm}
\centering
\caption{
 Performances (UP/LP accuracies) of the locked model using different locking methods on Pets dataset. R50 stands for Resnet50 and BadNet-$X$ represents BadNet trigger pattern $\text{patch size} = X$ 
} % \caption
\resizebox{0.9\textwidth}{!}{
\begin{tabular}{c|ccccccc}
    \toprule
        \multirow{2}{*}{\textbf{Pre-trained}}  & {BadNets-5} & {BadNets-20} & {BadNets-50} & {Blend} & {WaNet} & {Nashville} & \textbf{Modellock(ous)} \\ 
        & (UP$\uparrow$/LP{\color{Red}{$\downarrow$}}) & (UP$\uparrow$/LP{\color{Red}{$\downarrow$}})& (UP$\uparrow$/LP{\color{Red}{$\downarrow$}})& (UP$\uparrow$/LP{\color{Red}{$\downarrow$}})& (UP$\uparrow$/LP{\color{Red}{$\downarrow$}})& (UP$\uparrow$/LP{\color{Red}{$\downarrow$}})& (UP$\uparrow$/LP{\color{Red}{$\downarrow$}})\\
        \midrule
        \multirow{1}{*}{\textbf{MAE}} & {92.64/{92.64}} & {92.61/{92.70}} & {92.70/{90.52}} & {91.39/{\color{Red}{5.42}}} & {91.82/{87.57}} & {92.83/{\color{Red}{21.37}}} & {92.12/{\color{Red}{15.07}}}\\
        \midrule
        \multirow{1}{*}{\textbf{SimCLR-R50}} & {85.17/{85.17}} & {85.06/{84.71}} & {85.53/{83.07}} & {86.45/{\color{Red}{7.00}}} & {84.74/{\color{Red}{60.94}}} & {86.59/{\color{Red}{12.97}}} & {85.45/{\color{Red}{11.26}}} \\
    \bottomrule
    \end{tabular}
}
\label{tab:traditional_backdoor}
% \vspace{-0.1cm}
\end{table*}
\paragraph{Semanticless Key Prompts}
Besides the semantic key prompts studied above, we also explore a sequence of meaningless characters added to the image through the key prompt. Note that meaningless character key prompts {{like {"{\#}9T8oN07y1{\%}6{\$}"}}} incur only imperceptible modifications to the image, completely different from directly adding the characters to the image.
The result in Table S3 in supplementary materials suggests that while still viable, the lock's effectiveness is somewhat diminished compared to semantically meaningful prompts like \emph{styles} and \emph{objects}. {As further illustrated in Figure S1 in supplementary materials, the advantage of such prompts is their ability to generate highly inconspicuous edited images. Even in cases where training images have been leaked, it remains challenging for attackers to revert the prompts. Employing a sequence of meaningless characters as the \emph{key prompt} significantly expands the range of possible prompt choices. This can potentially be generalized to hybrid key prompts that employ combinations of different types of key prompts, leading to an infinite number of key prompt choices. Such possibility renders our \emph{ModelLock} practical in real-world applications.}

\paragraph{Backdoor Triggers}
We have also tested the use of \emph{traditional backdoor triggers} to lock the model. {These include the patch trigger of BadNets \cite{gu2017badnets}, the full image trigger of Blend \cite{chen2017targeted}, and the invisible, sample-wise triggers of WaNet \cite{nguyen2021wanet} and Nashville \cite{liu2019abs}. It is important to note that, to satisfy our proposed model lock paradigm, we need to create a ``poisoned" training dataset which consists of 1) 95\% backdoor samples with clean labels (by adding the trigger to the samples while keeping their labels unchanged), and 2) 5\% clean samples with modified labels (to a new and incorrect label). {{As demonstrated in Table \ref{tab:traditional_backdoor}, the full-image patterns and several sample-wise patterns can effectively lock the models. However, they were not selected as the primary methods for our Modellock due to insufficient security — an apparent full-image pattern such as "Hello Kitty" in the Blend lock or a fixed filter, that is, a function in the Nashville lock, can be easily exposed when the edited images are leaked. Moreover, it is worth mentioning that backdoor patterns are generally less flexible than our generative approach in customizing the key prompt for different users. This is so even for sample-wise backdoor patterns as they have a very limited number of tunable hyper-parameters for customization. The hyper-parameters can be easily recovered by grid search if the trigger type can be obtained. Specific experimental details and sample images can be found in Part C of the supplementary materials.}}

% We use `BadNets-5', `BadNets-20', and `BadNets-50' to denote trigger patch sizes of 5, 20, and 50, respectively.  For the Blend attack, the blending ratio is set to 8:2 (image:pattern) to ensure a more effective locking effect. For the WaNet attack, we slightly modified its setting into a data poison-only backdoor setting (following prior work \cite{huang2023distilling}) to guarantee the uniformity of the model training process. The grid size for WaNet is set to $k=32$.} Example images with backdoor patterns can be found in Figure S2 in supplementary materials.

% As demonstrated in Table \ref{tab:traditional_backdoor}, several full-image patterns and sample-wise patterns can effectively lock the models, while others are less effective. However, it is worth mentioning that backdoor patterns are generally less flexible than our generative approach in customizing the key prompt for different users. This is so even for sample-wise backdoor patterns as they have a very limited number of tunable hyper-parameters for customization. And the hyper-parameters can be easily recovered by grid search if the trigger type is exposed.

\begin{figure}[t]
    \centering
    \subfigure{
    \includegraphics[width=0.47\columnwidth]{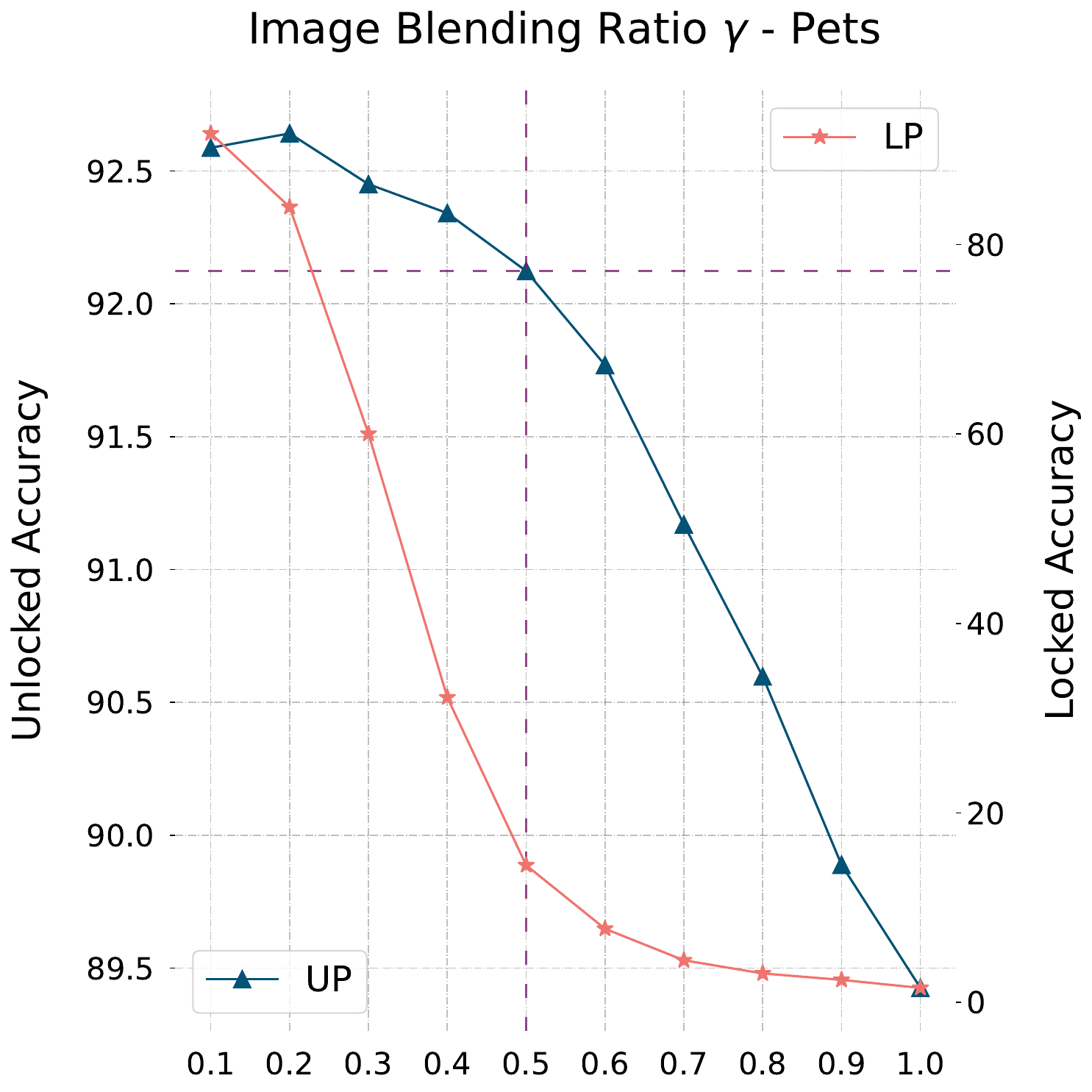}
    }
    \subfigure{
    \includegraphics[width=0.47\columnwidth]{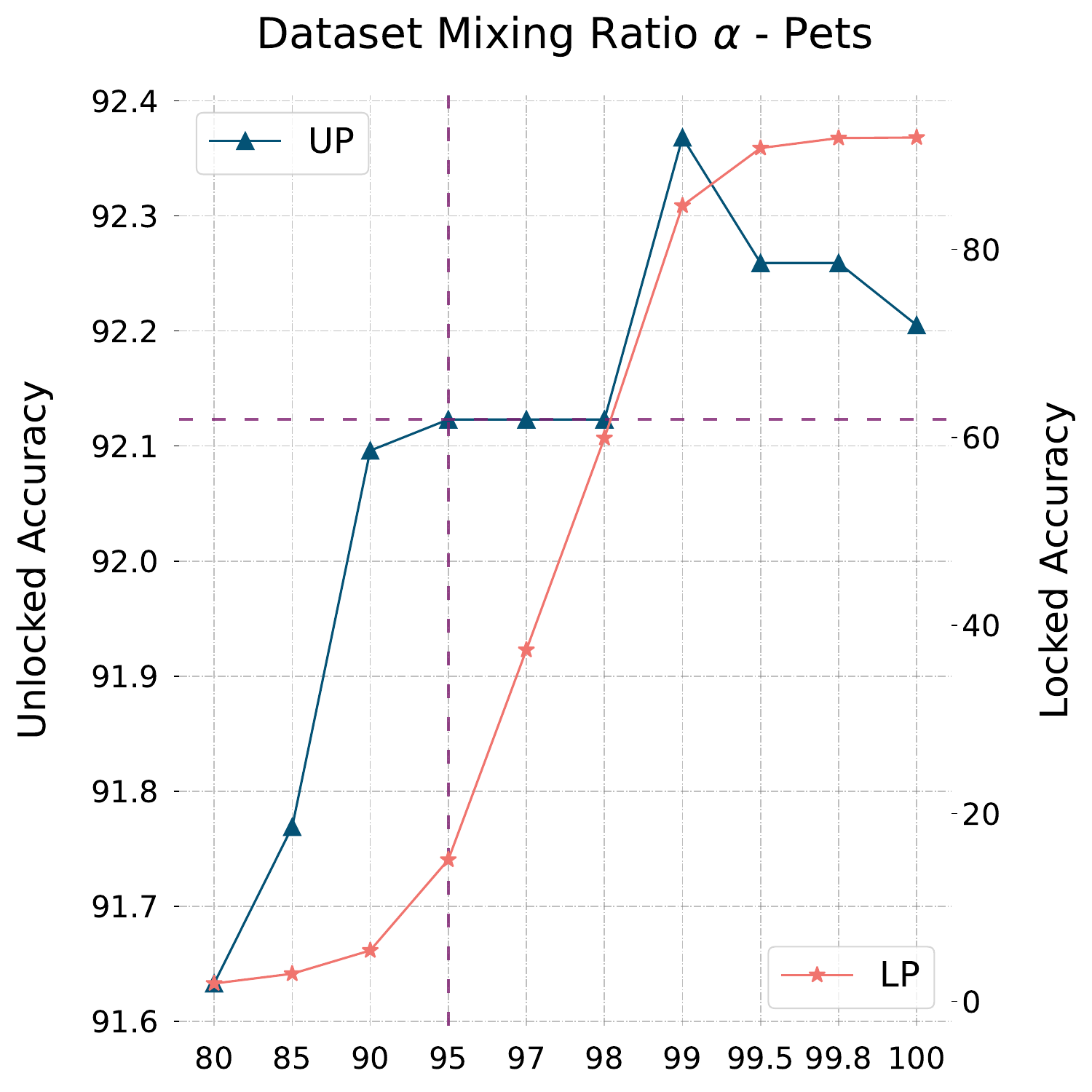}
    }
    \vspace{-0.2cm}
    \caption{The UP (left $y$-axis) and LP (right $y$-axis) performances of ModelLock under different hyperparameters. \textit{Left}: sensitivity to image blending ratio ($\gamma$); \textit{Right}: sensitivity to dataset mixing ratio ($\alpha$).}
    \label{plot:blening_&_mix_rate}
    \vspace{-0.3cm}

\end{figure}
\subsection{Ablation Studies}
We conduct ablation studies using the MAE-pretrained ViT-base model on the Pets dataset.

\paragraph{Image Blending Ratio $\gamma$} Despite the efforts to preserve the image's main subject when generating images using an image editing model, the manipulation of image semantics by the \textit{key prompt} still introduces noise that affects the fine-tuning process of the model. To address this issue, we adopt the popular image-blending technique used in backdoor attacks to reduce the impact of noise. Figure \ref{plot:blening_&_mix_rate} (left) shows that as the proportion of the edited image in the synthesized image increases, the LP significantly decreases compared to UP. 
% Specifically, the LP initially decreases rapidly and then gradually levels off, eventually no longer dropping. 
Therefore, selecting an appropriate image blending ratio can ensure the fine-tuned model remains locked while maximizing the UP on the edited dataset.

\paragraph{Dataset Mixing Ratio $\alpha$}  Figure \ref{plot:blening_&_mix_rate} (right) demonstrates that as the proportion of edited samples in the training set increases and the proportion of locked dataset decreases, the LP (lower is better) of the locked model rises rapidly, especially when the proportion of edited dataset reaches around 95$\%$. 

This indicates that the locking effect of the model deteriorates rapidly when the edited proportion exceeds a certain threshold (not much data left in the locking subset $\mathcal{D}_L$. Therefore, a trade-off is necessary when increasing the proportion of the edited dataset between the benefits brought by UP and the damage to the LP of the model.

\subsection{Robustness to Adaptive Attacks} \label{sec:Adaptive attacks}
Here, we test the robustness of \emph{ModelLock} to adaptive attacks.
We assume the attacker knows the open-source diffusion model and its detailed setup for ModelLock but not the \textit{key prompt}. We also assume the defender can download different LoRA-tuned weights from Hugging Face or Civitai to make the editing model more covert. That is, the attacker only knows the diffusion model but not the LoRA module.

Following the above assumption, we test two adaptive attack scenarios: (1) the edited images are not leaked, and (2) a few edited images are leaked. In case (1), the attacker cannot obtain the edited data, so it has to guess the key prompt. 
We call this attack \textit{Random Prompt (RP)}. In case (2), the attacker can access part of the edited data and use the data to estimate the key prompt with the help of GPT-4V. We call this attack \textit{Surrogate Prompt (SP)}. 
{{We are interested in the unlocked performance by the two attacks which are denoted as $\text{UP}_{rp}$ and $\text{UP}_{sp}$, respectively. 
Note that the use of LoRA by the defender may slightly decrease the model's task performance.}}

\paragraph{Unlocking Attack with \textit{Random Prompt}} We test a variety of \textit{random guessing prompts}, including adding random objects to the background, image style transfer, traditional image modification, empty prompts, and meaningless string prompts, to represent readily guessable prompts for the adversaries. 
As shown in Figure \ref{plot:random_prompt} (left), \emph{Style} lock demonstrates strong robustness against these attacks, while utilizing a customized diffusion model with LoRA can further enhance its robustness. The average $\text{UP}_{rp}$ of \emph{Style} lock is 62.82$\%$, which is still ~30$\%$ lower than that of using the correct \textit{key prompt} for unlocking. After incorporating LoRA weights, the average $\text{UP}_{rp}$ of \emph{Style} lock is further reduced to 6.46$\%$, rendering \textit{Random Prompt} attacks ineffective. 
However, as shown in Figure \ref{plot:random_prompt} (right), \emph{New Object} locks are relatively more vulnerable to adaptive attacks, which again can be largely mitigated by the use of LoRA.

\paragraph{Unlocking Attack with \textit{Surrogate Prompt}} \label{sec:Surrogate Prompt} We randomly select 10 images from the edited dataset as the "leaked data". The attacker shows these images to GPT-4V~\cite{yang2023dawn} and asks it to suggest \textit{surrogate key prompts} to unlock the model.  It is worth noting that images edited using the \emph{adding irrelevant new object} prompt do not have significant changes if the number of diffusion inference steps is small (e.g., "Basketball" prompt in Figure S1 in supplementary materials). Also, considering the huge number of possible object categories, it is generally hard for the attacker to approximate object key prompts even with the help of GPT-4V. However, the types of known styles are quite limited. 
As shown in Table S4 in supplementary materials, the unlocked performance $\text{UP}_{sp}$ by surrogate \emph{Style} prompt attacks is $68.48\%$ on average, which is still 20$\%$ (on average) lower than UP using the right \textit{key prompt} for unlocking. Moreover, when pre-trained LoRA weights are used, only 37.19$\%$ accuracy (on average) is unlocked by the attack. These experiments manifest the robustness of our \emph{ModelLock} to surrogate prompt attacks when the private data is not fully leaked or pre-trained LoRA weights are used.

\section{Discussion and Conclusion}
In this paper, we proposed a model locking technique \emph{ModelLock} to protect finetuned models on private data from being leaked by accident or extracted by potential adversaries. \emph{ModelLock} leverages text-guided image editing with publicly available diffusion models to edit a large proportion (e.g., 95\%) of the private data and relabel the rest of the clean samples to a wrong label. A model trained on the edited data will perform poorly on clean data and is only functional on edited test data guided by the same text prompt (called \emph{key prompt}). We tested the effectiveness of our ModelLock in locking classification, object detection, and segmentation models, and demonstrated its robustness to potential unlocking attacks. The models locked by our ModelLock will lose their capability on normal clean data and thus are naturally robust to adversarial attacks and model extraction attacks, as they could only attack a ``wrong" model. In this sense, ModelLock offers a higher level of security than model watermarking or fingerprinting techniques. 

% While showing promising results in different vision tasks, our \emph{ModelLock} is only a preliminary exploration of the model lock paradigm. There is still much room for improvement in terms of choosing the right key prompt or using optimization strategies to reduce the negative impact on the unlocked performance. We will also explore more fine-grained locking strategies that can release varying performance levels given different keys.

{{While showing promising results in different vision tasks, our \emph{ModelLock} is only a preliminary exploration of the model lock paradigm. The remaining challenges include how to select or optimize the \textit{key prompt} and how to apply it to tasks that are sensitive to input editing, such as detecting modification of real images. Promising future research directions include 1) Quantifying the distribution shift and subject semantic similarity to establish a threshold for candidate \textit{key prompts} selection based on the editing results; 2) Using optimized editing with task-related constraints, such as employing an originality-preserving constraint to maintain the original properties of the image, or even integrating the lock design and editing model optimization processes into the objective function, such as using the loss of clean data with unperturbed raw labels for gradient ascent to form a bilevel optimization framework. 
It is also interesting to explore more fine-grained locking strategies that can release varying performance levels given different keys.}}

\begin{figure}[h]
    \centering
    \subfigure{
    \includegraphics[width=0.47\columnwidth]{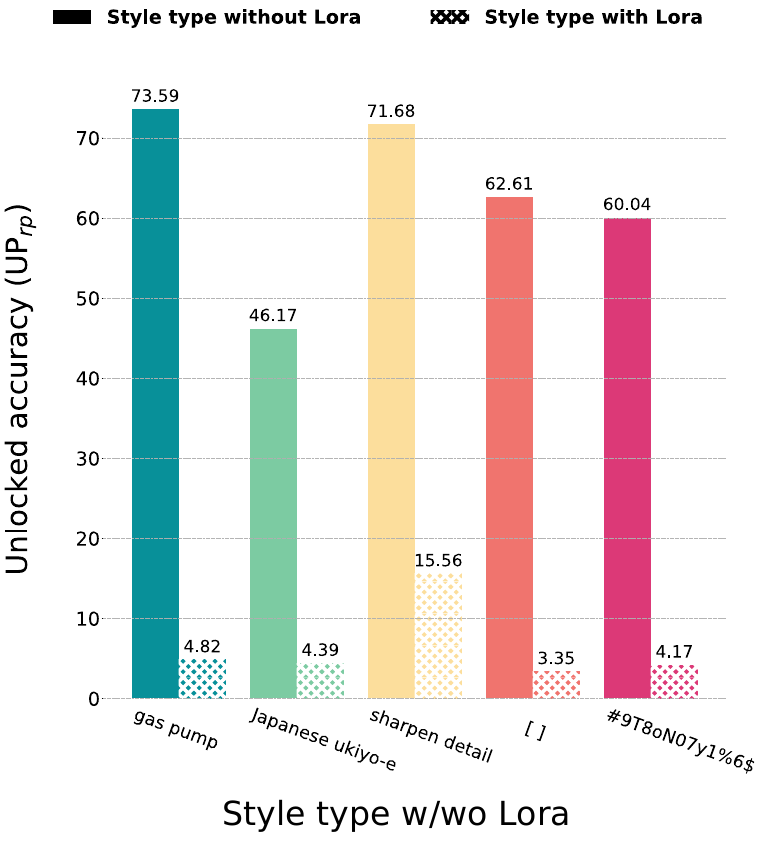}
    }
    \subfigure{
    \includegraphics[width=0.47\columnwidth]{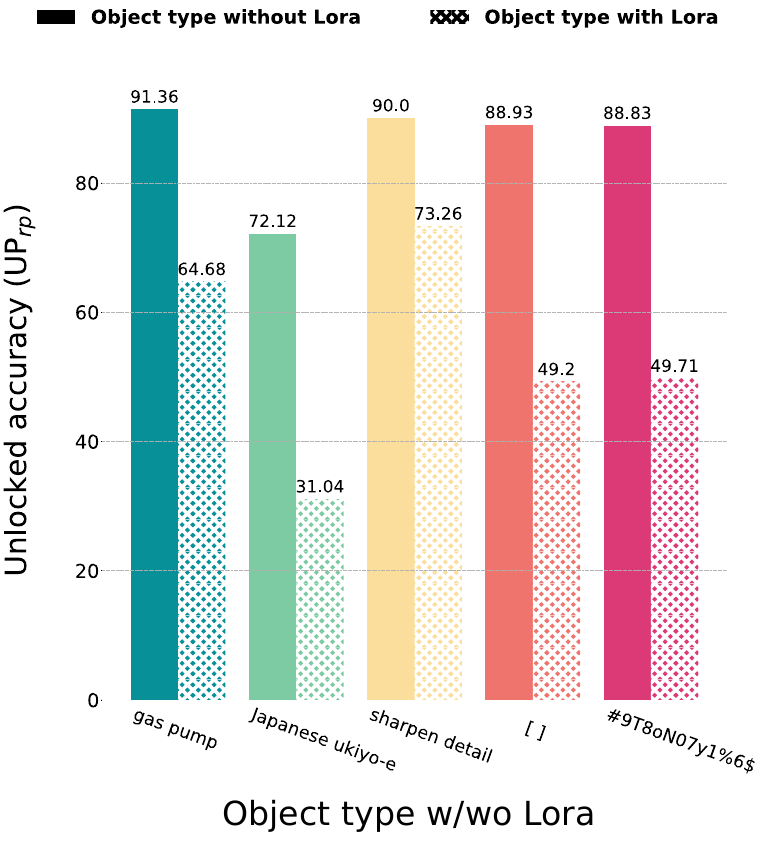}
    }
    \vspace{-0.5cm}
    \caption{Unlocked accuracy ($\text{UP}_{rp}$(\%)) by 5 types of \textit{random prompts} on Pets dataset. We test models locked using pre-trained Stable Diffusion with or without (w/wo) LoRA.}
    \label{plot:random_prompt}
    \vspace{-0.5cm}
\end{figure}


\begin{acks}
This work was supported by the National Natural Science Foundation of China under Grant No. 62276067.
\end{acks}

\bibliographystyle{ACM-Reference-Format}
\balance
\bibliography{sample-base}

\newpage
\appendix
\onecolumn
\setcounter{figure}{0}
\setcounter{table}{0}
\renewcommand{\thefigure}{S\arabic{figure}}
\renewcommand{\thetable}{S\arabic{table}}

\section{{Style} and {New Object} lock prompts in Section 4.3}
\label{sec:GPT4_prompt}
To evaluate the performance of different types of \textit{Style} and \textit{New Object} locks, here we utilize GPT4 to find 5 candidate prompts for each type of \textit{Style} lock and 25 candidate prompts for each type of \textit{New Object} lock. By exploring these types of locks, we aim to obtain practical recommendations more comprehensively. However, it is important to acknowledge that there exist many possible alternatives for \textit{Style} and \textit{New Object} lock prompts, and our exploration may have limitations.

% \begin{table}[h]
% \vspace{-0.2cm}
% \centering
% \resizebox{0.98\linewidth}{!}{
% \begin{tabular}{l|c|c|c}
%     \toprule[2pt]
%         \multirow{1}{*}{\textbf{Painting Styles}} & \textbf{Artist Styles} & \textbf{Art Forms} & \multirow{1}{*}{\textbf{Traditional Photo Adjustment}}   \\ 
%         \midrule[1pt]
%         \text{Watercolor painting effect} & \text{With Van Gogh's painting style} & \text{With cyberpunk style} & \text{Boost color saturation}  \\
        
%         \text{Impasto brush strokes} & \text{In Monet's impressionist style} & \text{Art deco geometric patterns} & \text{Sharpen image detail}  \\
        
%         \text{Acrylic vibrant colors} & \text{Rembrandt lighting technique} & \text{Pop art comic book flair} & \text{Balance contrast levels}  \\
        
%         \text{Charcoal sketch texture} & \text{Picasso's cubist interpretation} & \text{Minimalist Scandinavian design} & \text{Adjust white balance for warmth}  \\
        
%         \text{Pastel tones overlay} & \text{Dalí's surrealism touch} & \text{Japanese ukiyo-e woodblock prin} & \text{Enhance shadows and highlights}  \\
%     \bottomrule[2pt]
%     \end{tabular}
% }
% \caption{Candidate prompts from GPT4 for testing different kinds of \textit{Style} lock \emph{ModelLock}'s Unlocked Performance (UP) and Locked Performance (LP).
% % 

% }
% \label{tab:Style_key_prompt}
% % \vspace{-0.5cm}
% \end{table}
\begin{table*}[!h]
% \vspace{-0.2cm}
\centering
\caption{
Candidate prompts from GPT4-V for testing different types of \textit{Style} locks in terms of the Unlocked Performance (UP) and Locked Performance (LP) of \emph{ModelLock}.}
\resizebox{0.95\textwidth}{!}{
\begin{tabular}{c|c|c|c}
    \toprule
        \multirow{1}{*}{\textbf{Painting Styles}} & \textbf{Artist Styles} & \textbf{Art Forms} & \multirow{1}{*}{\textbf{Traditional Photo Adjustment}}   \\ 
        \midrule
        \text{Watercolor painting effect} & \text{With Van Gogh's painting style} & \text{With cyberpunk style} & \text{Boost color saturation}  \\
        
        \text{Impasto brush strokes} & \text{In Monet's impressionist style} & \text{Art deco geometric patterns} & \text{Sharpen image detail}  \\
        
        \text{Acrylic vibrant colors} & \text{Rembrandt lighting technique} & \text{Pop art comic book flair} & \text{Balance contrast levels}  \\
        
        \text{Charcoal sketch texture} & \text{Picasso's cubist interpretation} & \text{Minimalist Scandinavian design} & \text{Adjust white balance for warmth}  \\
        
        \text{Pastel tones overlay} & \text{Dalí's surrealism touch} & \text{Japanese ukiyo-e woodblock prin} & \text{Enhance shadows and highlights}  \\
    \bottomrule
\end{tabular}

}
% 
% \caption
\label{tab:Style_key_prompt}
% \vspace{-0.6cm}
\end{table*}
When generating \emph{Irrelevant Object} and \emph{Relevant Object} with GPT4-V, we provide a list of labels of ImageNet-1K as a reference. In Table \ref{tab:Object_key_prompt}, the parentheses following each returned \emph{candidate object prompt} indicate the corresponding ImageNet-1K label index. Interestingly, we find that GPT4-V may produce \emph{object prompts} that are not in ImageNet-1K (e.g., seahorse), suggesting that the range of potential \emph{key prompts} is quite extensive.

\begin{table*}[!]
% \vspace{-0.2cm}
\centering
\caption{
Candidate prompts from GPT4-V for testing different types of relevant and irrelevant \textit{Object} locks in terms of the Unlocked Performance (UP) and Locked Performance (LP) of \emph{ModelLock}.}
\resizebox{0.95\textwidth}{!}{
\begin{tabular}{c|c}
    \toprule[2pt]
Irrelevant Object & Relevant Object  \\ \hline
1. abacus (398) & 1. great white shark, white shark, man-eater, man-eating shark, Carcharodon carcharias (2) \\ \hline
2. accordion, piano accordion, squeeze box ( 401 ) & 2. bald eagle, American eagle, Haliaeetus leucocephalus (22)                                \\ \hline
3. aircraft carrier, carrier, flattop, attack aircraft carrier (403) & 3. African elephant, Loxodonta africana (386) \\ \hline
4. analog clock (409) & 4. chimpanzee, chimp, Pan troglodytes (367) \\ \hline
5. apiary, bee house (410)  & 5. gorilla, Gorilla gorilla (366) \\ \hline
6. barometer (426) & 6. koala, koala bear, kangaroo bear, native bear, Phascolarctos cinereus (105) \\ \hline
7. baseball (429) & 7. lion, king of beasts, Panthera leo (291) \\ \hline
8. basketball (430) & 8. tiger, Panthera tigris (292) \\ \hline
9. bathtub, bathing tub, bath, tub (435) & 9. cheetah, chetah, Acinonyx jubatus (293) \\ \hline
10. beer bottle (440) & 10. hippopotamus, hippo, river horse, Hippopotamus amphibius (344) \\ \hline
11. bow tie, bow-tie, bowtie (457) & 11. giraffe (Giraffa camelopardalis) (not on the list but a relevant animal not similar to the subject) \\ \hline
12. bullet train, bullet (466) & 12. zebra (340) \\ \hline
13. cello, violoncello (486) & 13. African grey, African gray, Psittacus erithacus (87) \\ \hline
14. chain saw, chainsaw (491) & 14. peacock (84) \\ \hline
15. church, church building (497) & 15. flamingo (130) \\ \hline
16. corkscrew, bottle screw (512) & 16. jellyfish (107) \\ \hline
17. digital clock (530) & 17. sea urchin (328) \\ \hline
18. electric fan, blower (545) & 18. starfish, sea star (327) \\ \hline
19. fire engine, fire truck (555) & 19. kangaroo (not listed but a relevant animal not similar to the subject) \\ \hline
20. flute, transverse flute (558) & 20. orangutan, orang, orangutang, Pongo pygmaeus (365) \\ \hline
21. gas pump, gasoline pump, petrol pump, island dispenser (571) & 21. giant panda, panda, panda bear, coon bear, Ailuropoda melanoleuca (388)  \\ \hline
22. grocery store, grocery, food market, market (582) & 22. rhinoceros (not listed but a relevant animal not similar to the subject) \\ \hline
23. hair slide (584) & 23. seahorse (not listed but a relevant animal not similar to the subject) \\ \hline
24. mailbox, letter box (637) & 24. octopus (not listed but a relevant animal not similar to the subject) \\ \hline
25. parachute, chute (701) & 25. coral reef (973) \\
    \bottomrule[2pt]
\end{tabular}

}
% 
% \caption
\label{tab:Object_key_prompt}
% \vspace{-0.6cm}
\end{table*}

\begin{table*}
% \vspace{-0.2cm}
\centering
\caption{
 Performances (UP/LP accuracies) of the Modellock locked model (using combinations of chrachters devoid of semantic content as key prompt) on Pets dataset. R50 stands for Resnet50.
} % \caption
\resizebox{0.55\linewidth}{!}{
\begin{tabular}{c|c|c}
    \toprule
        \multirow{2}{*}{\textbf{Pre-trained}} & \multirow{2}{*}{\textbf{Meaningless Prompt}} & \textbf{Pets} \\ 
        && (UP$\uparrow$/LP{\color{Red}{$\downarrow$}}) \\
        \midrule
        \multirow{2}{*}{\textbf{MAE}} & {with \text{'@de\%\^{}fwd{\#}oi'} in the background} & {91.55/{\color{red}{52.74}}} \\
        & {{\#}9T8oN07y1\%6\$} & {91.66/{\color{red}{58.60}}} \\
        \midrule
        \multirow{2}{*}{\textbf{SimCLR-R50}} & {with '@de\%\^{}fwd{\#}oi' in the background} & {84.63/{\color{Red}{32.03}}} \\
        & {{\#}9T8oN07y1{\%}6{\$}} & {84.03/{\color{red}{25.16}}} \\
    \bottomrule
    \end{tabular}
}
\label{tab:chrachters}
% \vspace{-0.5cm}
\end{table*}

% \begin{table}
% % \vspace{-0.2cm}
% \centering
% \resizebox{0.45\textwidth}{!}{
% \begin{tabular}{l|c|c}
%     \toprule[2pt]
%         \multirow{2}{*}{\textbf{Pre-trained}} & \multirow{2}{*}{\textbf{Meaningless Prompt}} & \textbf{Pets} \\ 
%         && (UP$\uparrow$/LP{\color{Red}{$\downarrow$}}) \\
%         \midrule[1pt]
%         \multirow{2}{*}{\textbf{MAE}} & {with '@de\%\^{}fwd{\#}oi' in the background} & {91.55/{\color{red}{52.74}}} \\
%         & {{\#}9T8oN07y1\%6\$} & {91.66/{\color{red}{58.60}}}
%         \midrule[1pt]
%         \multirow{1}{*}{\textbf{SimCLR-R50}} & {with '@de\%\^{}fwd{\#}oi' in the background} & {84.63/{\color{Red}{32.03}}} \\
%         & {{\#}9T8oN07y1{\%}6{\$}} & {84.03/{\color{red}{25.16}}}

%     \bottomrule[2pt]
%     \end{tabular}
% }
% \caption{
% % 
% Whether current model protection methods, such as Watermarking, Fingerprinting, and our newly proposed ModelLock, can be effective in model verification and model deactivation.
% } % \caption
% \label{tab:sign}
% \vspace{-0.5cm}
% \end{table}
\begin{table*}[!t]
% \vspace{-0.2cm}
\centering
\caption{
A list of surrogate prompts (generated by GPT-4V based on the leaked edited images) and their unlocking performance ($\text{UP}_{sp}(\%)$). 
The experiments were conducted with MAE-pretrained ViT-base and the Pets dataset.
} % \caption
\resizebox{\linewidth}{!}{
\begin{tabular}{l|c|c}
    \toprule
        \multirow{1}{*}{\textbf{Victim Model}} & \multirow{1}{*}{\textbf{Surrogate Prompts}} & \textbf{$\text{UP}_{sp}(\%)$}  \\ 
        \midrule
        \multirow{5}{*}{\textbf{Style lock \textbf{wo} Lora}} & {\text{Enhance the texture and fur detail of a domestic cat/dog while maintaining a natural look.}} & 60.53 \\
        & {\text{Apply a soft, dreamy filter to a pet portrait for a whimsical effect.}} & 79.31 \\
        & {\text{Increase the sharpness and clarity of a pet\'s facial features.}} & 63.21 \\
        & {\text{Adjust the lighting and saturation to create a vibrant, lively scene with a pet.}} & 68.03 \\
        & {\text{Create a painting-like effect to give a pet photo an artistic touch.}} & 71.30 \\
        \midrule
        \multirow{5}{*}{\textbf{Style lock \textbf{w} Lora}} & {\text{Dreamlike texture overlay on pet photos.}} & 30.25 \\
        & {\text{Enhanced saturation with a surreal touch for animal portraits.}} & \textbf{7.93}  \\
        & {\text{Oil painting effect applied to dogs and cats photography.}} & \textbf{7.69}  \\
        & {\text{Abstract color infusion for a whimsical look in pet images.}} & 61.30  \\
        & {\text{Pastel and neon color blend for a magical ambiance in animal pictures.}} & 78.79  \\
    \bottomrule
    \end{tabular}
}
\label{tab:Surrogate_Prompt_Attack}
% \vspace{13cm}
\end{table*}

\section{Example edited images with different prompts}\label{sec:Example_fig}
Here, we show example images from various image editing datasets. The examples include (1) \emph{style lock} prompts such as Painting Styles - "with oil paste"; Traditional Photo Adjustment - "Sharpen image detail"; Artist Styles - "With Van Gogh’s painting style" and Art Forms - "With cyberpunk style". (2) \emph{object lock} prompts such as Irrelevant Object - "with basketball in the background" and Relevant Object - "with Koala in the background". (3) \emph{meaningless lock}  prompt "\#9T8oN07y1\%6\$". All edited images are synthesized with a blending ratio $\gamma = 0.5$ on the raw images.

\renewcommand{\dblfloatpagefraction}{.9}
\begin{figure*}[t]
% \vspace{-0.4cm}
    \centerline{\includegraphics[width=0.85\textwidth]{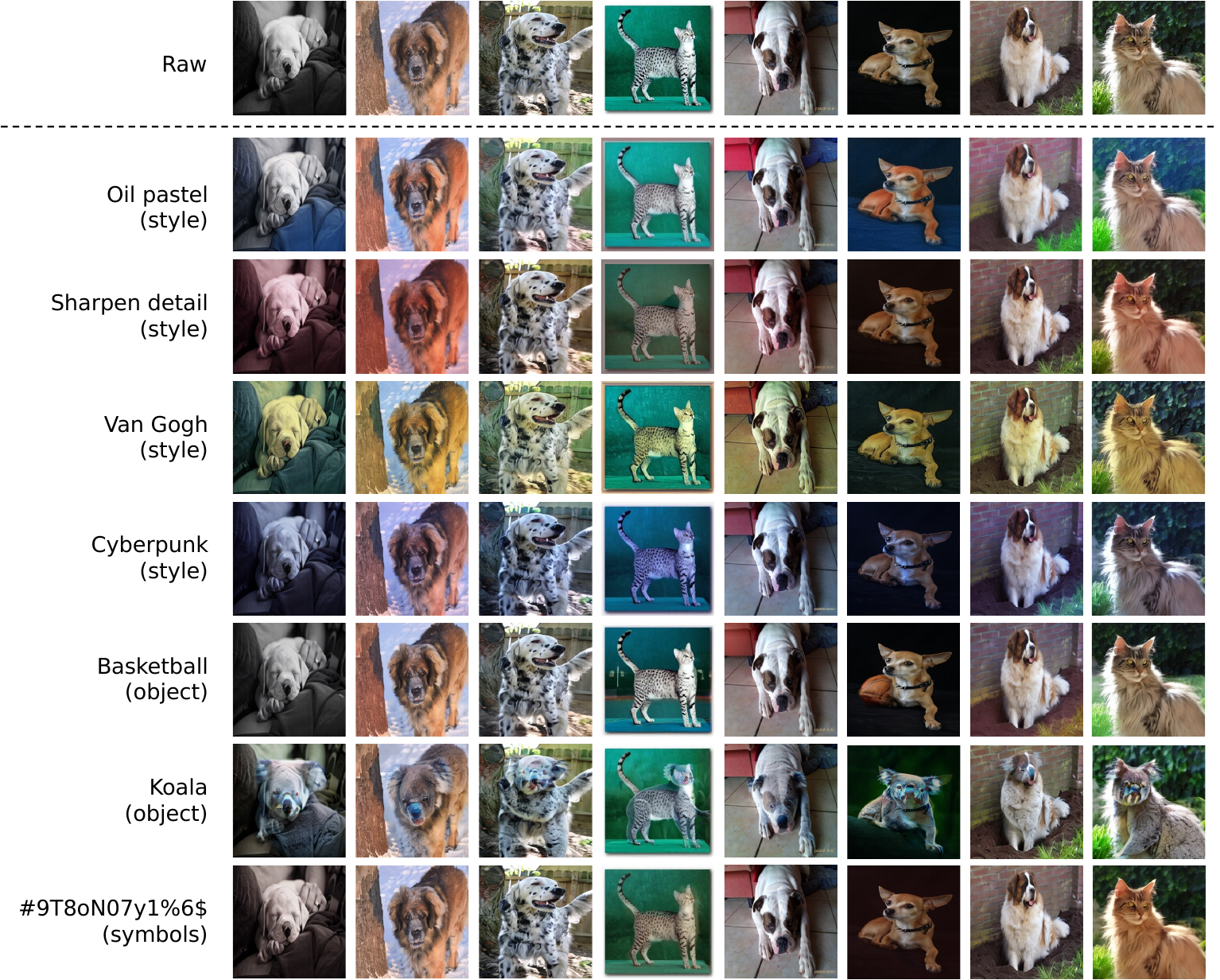}}
\caption{
Examples of edited images by \emph{Style}, \emph{Object}, and \emph{Combination of symbols} lock prompts. All examples shown here are edited images synthesized with blending ratio $\gamma = 0.5$ on the raw images after the diffusion editing process.
}
\label{fig:Show_fig}
\vspace{-0.1cm}
\end{figure*}

As can be observed in Figure \ref{fig:Show_fig}, the edited images exhibit certain discrepancies compared to the raw images, although these differences are not pronounced. Particularly for edited images using prompts such as "adding irrelevant object in the background (e.g., Basketball)" and "using a sequence of characters (e.g., "\#9T8oN07y1\%6\$")", the alterations are nearly imperceptible. One reason for this is that we employed significantly fewer inference steps (e.g., inference step = 5) than typical image editing. This has two advantages: 1) it improves the stealthiness of \emph{ModelLock}, and 2) it reduces the computational cost of image editing.

Moreover, the edited images using prompts such as "adding relevant object in the background (e.g., Koala)" exhibit more obvious alterations to the raw images and sometimes can affect the main objects. This tends to introduce new noise into the edited images, impairing the model's performance (UP). Therefore, in Section 3.3, we made \textit{practical recommendations} to use "adding unrelated new object in the background".

\section{Experimental setup and example images of the traditional backdoor lock}
We use `BadNets-5', `BadNets-20', and `BadNets-50' to denote trigger patch sizes of 5, 20, and 50, respectively.  For the Blend attack, the blending ratio is set to 8:2 (image:pattern) to ensure a more effective locking effect. For the WaNet attack, we slightly modified its setting into a data poison-only backdoor setting (following prior work \cite{huang2023distilling}) to guarantee the uniformity of the model training process. The grid size for WaNet is set to $k=32$.

As shown in Figure \ref{fig:traditional_bd}, the patch pattern of BadNets and the predetermined pattern "Hello Kitty" of Blend cause the edited samples to be distinguishable from the raw images. In the adaptive attack scenarios described in Section 4.5, when a few edited images are leaked, these patterns can be easily perceived and exploited by the attacker to unlock the model, undermining the security of \emph{ModelLock}.
However, the modifications by our \emph{ModelLock} are minimized by the diffusion model, rendering them virtually imperceptible. Arguably, without the raw images, the edited images by \emph{ModelLock} do not stand out conspicuously.
\begin{figure*}[h]
% \vspace{-0.4cm}
\centerline{\includegraphics[width=0.85\textwidth]{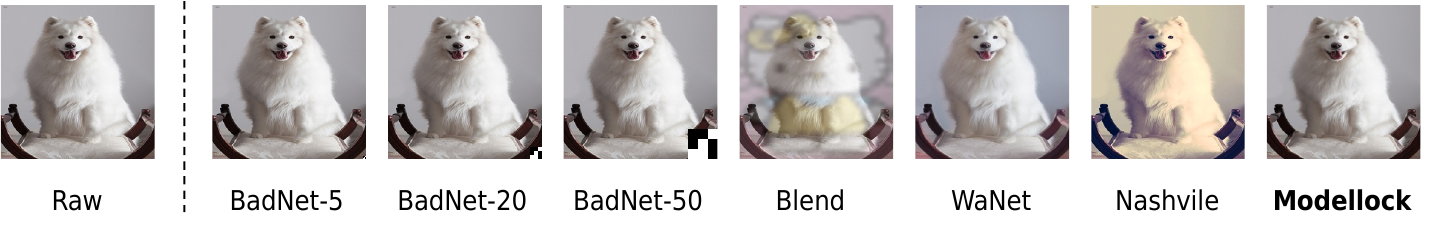}}
\vspace{-10pt}
\caption{
Example edited images using traditional backdoor patterns (BadNets and Blend) as locks. Note that \text{BadNets-5}, \text{BadNets-20}, and \text{BadNets-50} correspond to patch trigger sizes of 5, 20, and 50, respectively.
}
\label{fig:traditional_bd}
\vspace{-0.1cm}
\end{figure*}

\section{Several questions about Modellock}
\quad{\,}{\textbf{Q1}. }\textbf{What are the differences between Modellock and alternative model protecting approaches?}

Model lock is necessary when fine-tuning private models on Machine Learning as a Service (MLaaS) platforms to protect the model and training data from dishonest service providers. The private model/data owner can utilize a customized LoRA-equipped diffusion model to pre-process the data before uploading it to the MLaaS platform. And the trained (locked) model is unusable even if the service provider leaks it to the model/data owner's competitors. It is also important to note that model encryption and model lock are two distinct types of model protection methods, and our ModelLock is compatible with model encryption.\\

{\textbf{Q2}. }\textbf{Can the key be stolen?}

It is difficult to reverse-engineer the key without the (clean - edited) image pairs due to the large search space of text tokens. Additionally, even with the
correct key prompt, the model cannot be unlocked without knowing
the exact diffusion model used (equipped with customized LoRA weights). Nonetheless, more strong key prompt attacks such as Monte
Carlo optimization or repeated attack sequences being optimized should be further explored on our ModelLock.\\

{\textbf{Q3}. }\textbf{Can Modellock be applied to LLM and MLM?}

Despite LLMs' superiority in general domains, large models perform only mediocrely in specialized and private domains (e.g., medical, financial, self-driving, embodied AI) due to a lack of training data. Therefore, for LLMs, ModelLock aims to capitalize on the lack of generalizability in these specialized domains to lock proprietary LLMs. Therefore, Modellock is still effective for LLM and MLM.

\begin{figure}[t]
    \centering
    \subfigure{
    \includegraphics[width=0.36\columnwidth]{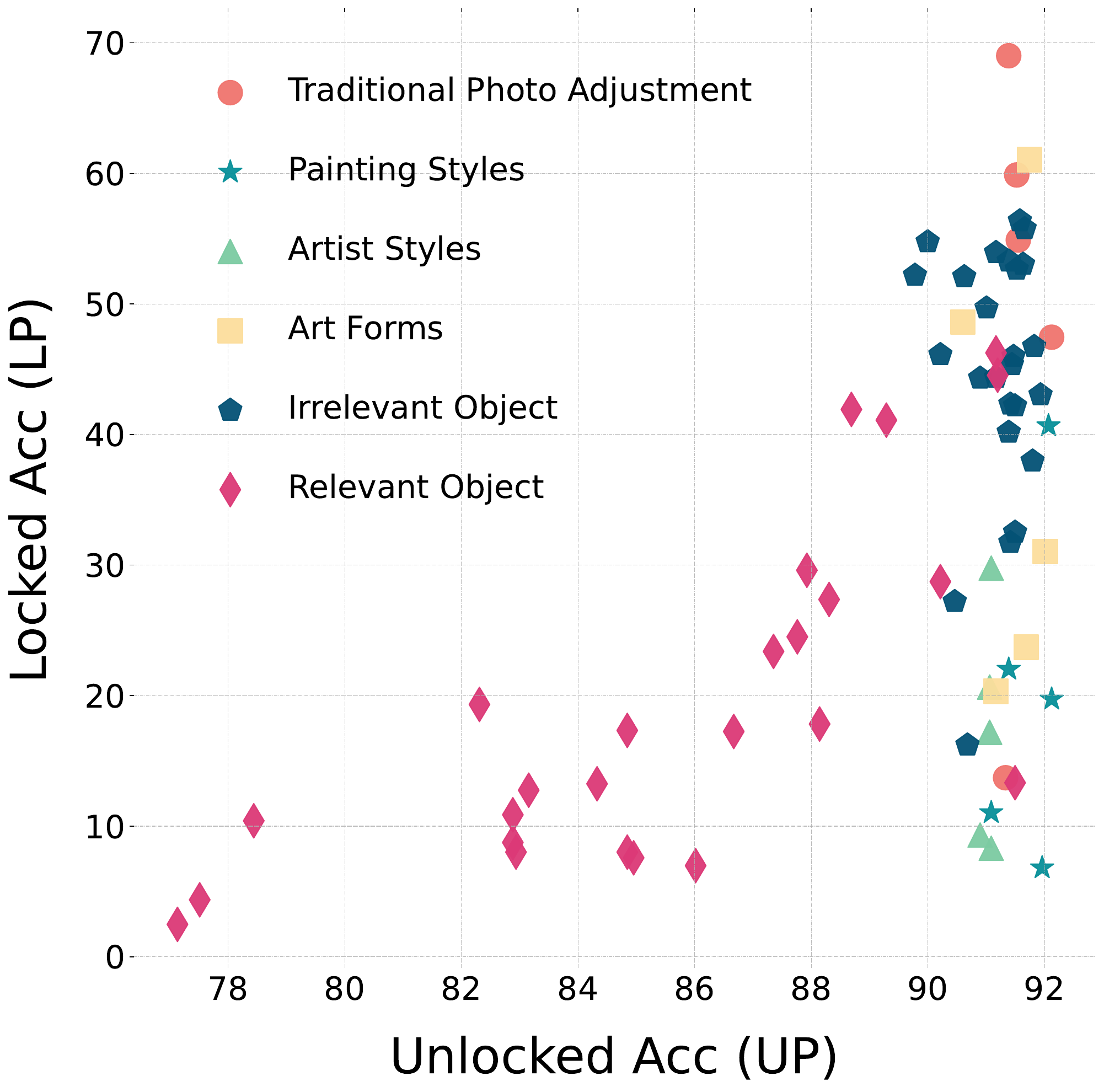}
    }
    \subfigure{
    \includegraphics[width=0.36\columnwidth]{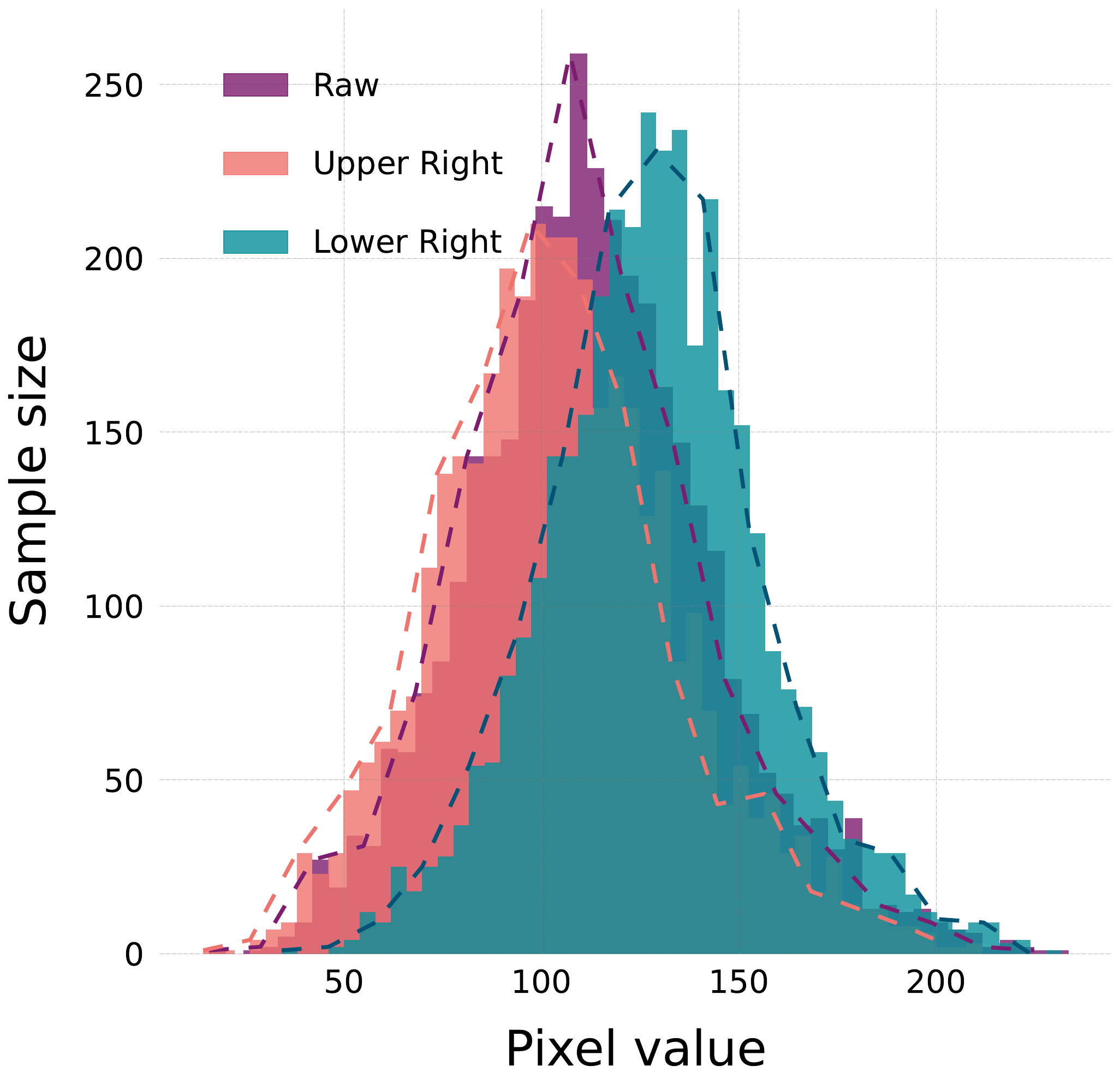}
    }
    \vspace{-13pt}
    \caption{\textbf{Left}: Unlocked accuracy (UP) and locked accuracy (LP) of models fine-tuned on the Pets dataset using 70 different \textit{key prompts} of 6 different types. \textbf{Right}: Pixel value distributions of the raw (purple) and edited (pink vs. green) datasets, with green editing are better model locks.}
    \label{plot:scatter_density}
    \vspace{-15.5pt}
\end{figure}

\end{document}